\documentclass{bmvc2k}

\title{Probing Association Instability with Track-State Perturbations for Clip-Level Active Learning in Query-Propagation Multi-Object Tracking}

\addauthor{Riku Inoue}{riku.inoue@ntt.com}{1}
\addauthor{Shogo Sato}{shg.sato@ntt.com}{1}
\addauthor{Kazuhiko Murasaki}{kazuhiko.murasaki@ntt.com}{1}
\addauthor{Tomoyasu Shimada}{tomoyasu.shimada@ntt.com}{1}
\addauthor{Toshihiko Nishimura}{toshihiko.nishimura@ntt.com}{1}
\addauthor{Ryuichi Tanida}{ryuichi.tanida@ntt.com}{1}

\addinstitution{
 Human Informatics Laboratories\\
 NTT, Inc.\\
 Kanagawa, Japan
}

\runninghead{INOUE ET AL.}{QPID for Clip-Level Active Learning in MOT}

\usepackage{graphicx}
\usepackage{booktabs}
\usepackage{adjustbox}
\usepackage{capt-of}
\usepackage{amsmath}
\usepackage{amssymb}

\begin{document}

\maketitle

\begin{abstract}
Training query-propagation end-to-end multi-object tracking (MOT) models requires dense bounding-box and identity annotations across video sequences, making dataset construction expensive.
Clip-level active learning reduces this cost by selecting video clips for annotation, but prior acquisition criteria based on output-level temporal uncertainty may miss clips whose informativeness comes from association instability in propagated track states.
We propose \textbf{QPID} (\textbf{Q}uery-\textbf{P}ropagation \textbf{I}nstability and \textbf{D}iversity), a clip acquisition method for query-propagation MOT that targets association instability in propagated track states.
QPID estimates this instability by applying two-sided perturbations to internal track states and measuring prediction differences from a clean reference branch.
The key idea is that, in stable clips, each propagated track should continue to follow the same target under small perturbations, whereas in ambiguous clips, small changes in the track state can alter which target the track follows, leading to changes in localization or confidence.
QPID measures these perturbation-induced prediction differences with two metrics: \textit{Localization Drift} and \textit{Entropy-Weighted Confidence Discrepancy}.
These metrics are aggregated into a clip-level association-instability score.
To avoid redundant uncertainty-only selection, QPID selects a representative annotation batch from high-instability clips using \textit{Uncertainty-Weighted Visual Coverage} with track-level visual prototypes.
Experiments on DanceTrack and SportsMOT with MeMOTR and SambaMOTR show that QPID achieves strong performance compared with active learning baselines under the same annotation budget.
\end{abstract}
 
\begin{figure*}
\centering
  \includegraphics[width=0.95\textwidth]{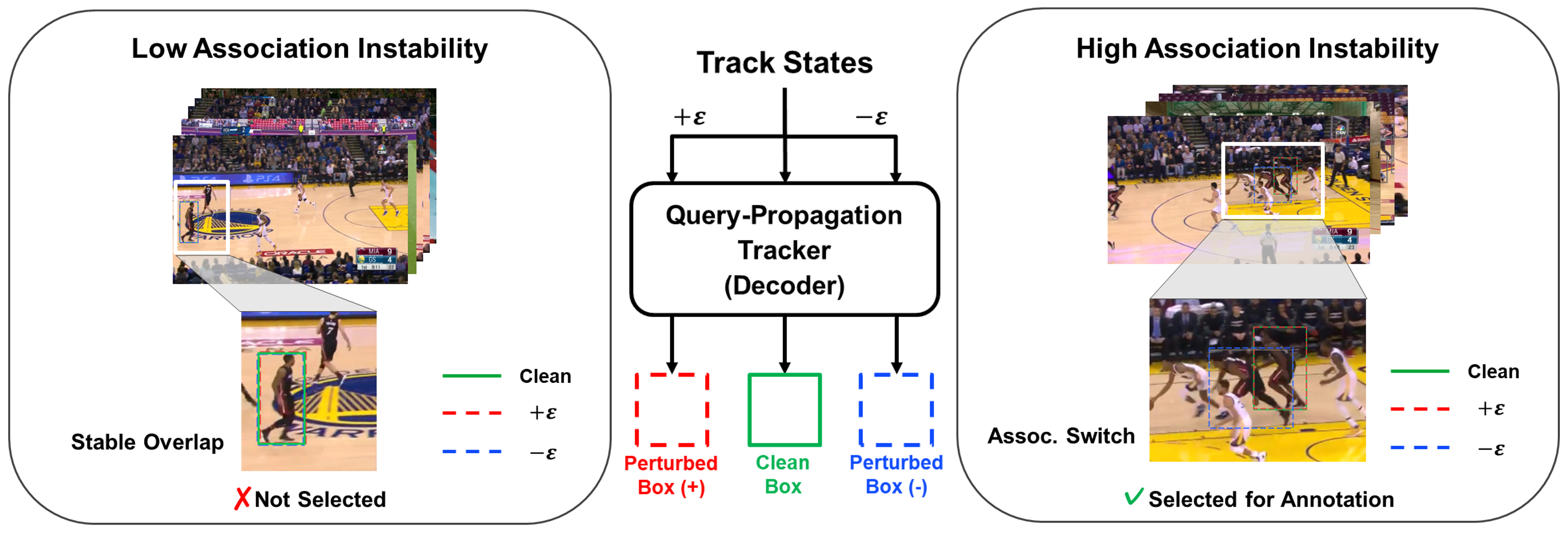}
    \caption{\textbf{Association instability via two-sided perturbations to internal track states.} 
    QPID applies small two-sided perturbations ($\pm\varepsilon$) to internal track states, consisting of track query embeddings and reference points, and compares the resulting predictions with a clean branch.
    Stable overlap between the clean and perturbed boxes indicates low association instability, whereas localization drift or association switches indicate high association instability.
    Green denotes the clean box, and red and blue dashed boxes denote the perturbed boxes for the track prediction visualized in each example.}
  \label{fig:fig1}
\end{figure*}
\section{Introduction}
\label{sec:intro}

Multi-Object Tracking (MOT) aims to localize multiple objects in a video and maintain their identities over time.
For many years, MOT has been dominated by the tracking-by-detection paradigm~\cite{zhang2022bytetrack,cao2023observation,aharon2022bot}, where objects are detected in each frame and associated across frames using motion or appearance cues.
While this paradigm is highly effective on conventional pedestrian benchmarks~\cite{milan2016mot16,dendorfer2020mot20}, it can struggle in benchmarks with irregular motion and many appearance-similar targets, such as DanceTrack~\cite{sun2022dancetrack} and SportsMOT~\cite{cui2023sportsmot}.
End-to-end trackers have therefore emerged as a strong direction by maintaining identities through propagated track queries~\cite{zeng2022motr,meinhardt2022trackformer,gao2023memotr,segu2025samba}.
These trackers have substantially improved tracking performance in such challenging scenes by learning to maintain identities among multiple plausible association candidates.

Despite these advances, training such trackers remains annotation-intensive.
MOT supervision requires dense bounding boxes together with temporally consistent identity labels across video sequences.
This cost becomes especially severe in crowded scenes, where annotators must resolve ambiguous associations over time.
Moreover, videos contain strong temporal redundancy, so naive annotation sampling can waste the budget on near-duplicate temporal segments.
Therefore, under a limited annotation budget, it is important to select clips that expose difficult association cases and are informative for improving identity maintenance in the target tracker.

Active Learning (AL) reduces annotation cost by prioritizing informative samples for labeling.
In computer vision, AL has been widely studied for image classification~\cite{sener2017active,citovsky2021batch,wang2016cost} and object detection~\cite{kao2018localization,yu2022consistency,yang2024plug}, where acquisition criteria are often based on uncertainty, representativeness, or their combination.
Video AL~\cite{rana2022all,goswami2023active,rana2023hybrid,singh2024semi} and MOT-oriented AL~\cite{li2023heterogeneous} have also explored sample selection under limited annotation budgets, often through frame-level acquisition.
However, selecting individual frames is not fully aligned with end-to-end trackers, whose training samples are multi-frame clips with temporally consistent identities.
Since identity maintenance in these trackers relies on temporal propagation across frames, evidence of association difficulty can be distributed across multiple frames within a clip rather than captured by a single frame.
Thus, acquisition should be performed at the clip level, allowing such evidence to be aggregated across multiple frames.

CUTAL~\cite{inoue2026clip} addresses the acquisition-unit mismatch by formulating clip-level active learning for end-to-end MOT.
By selecting fixed-length clips, CUTAL better aligns acquisition with the multi-frame training structure of end-to-end trackers than frame-level acquisition.
However, its output-level temporal uncertainty mainly reflects variations in final predictions, such as entropy variation or confidence inconsistency, and does not directly probe instability in propagated track states.
In query-propagation trackers, each propagated track state carries information about the target followed by the track across frames.
When appearance-similar targets approach or cross, final predictions may remain confident and temporally smooth, while small changes in the propagated state can change localization or alter which target the track follows.
Such instability exposes clips where the tracker must learn to maintain identities among competing targets, which may not be captured by output-level temporal uncertainty alone.

In this work, we propose \textbf{QPID}, \textbf{Q}uery-\textbf{P}ropagation \textbf{I}nstability and \textbf{D}iversity, a clip-level active learning framework for query-propagation MOT.
QPID estimates association instability by applying two-sided perturbations to internal track states and measuring prediction differences from a clean reference branch.
As illustrated in Fig.~\ref{fig:fig1}, in stable clips, perturbed predictions remain consistent with the clean prediction under small perturbations, whereas in ambiguous clips, a perturbed track state can lead to localization drift or an association change.
QPID aggregates these perturbation-induced prediction differences using \textit{Localization Drift} and \textit{Entropy-Weighted Confidence Discrepancy} into a clip-level association-instability score.
To avoid selecting redundant high-instability clips, QPID selects a representative annotation batch using \textit{Uncertainty-Weighted Visual Coverage} with track-level visual prototypes.

Our contributions are as follows.
\begin{itemize}
\vspace{-1mm}
\item We propose QPID, a clip-level active learning framework for query-propagation MOT that selects informative and non-redundant clips by estimating association instability in propagated track states and improving visual coverage.
\vspace{-1mm}
\item We introduce a two-sided perturbation scheme that applies paired perturbations to internal track states and quantifies perturbation-induced association instability using Localization Drift and Entropy-Weighted Confidence Discrepancy.
\vspace{-1mm}
\item We demonstrate the effectiveness of QPID on DanceTrack and SportsMOT with MeMOTR and SambaMOTR, showing strong performance against active learning baselines under the same annotation budget.
\vspace{-1mm}
\end{itemize}
\section{Related Work}
\label{sec_related}

\subsection{Multi-Object Tracking}
\label{sec_related_mot}

MOT has long been studied under the tracking-by-detection paradigm, where objects are detected in each frame and then associated over time.
Early studies were largely centered on pedestrian benchmarks~\cite{milan2016mot16,dendorfer2020mot20}, where pipelines combining strong detectors with post-hoc association became standard.
These methods typically combine object detectors~\cite{ge2021yolox,zhou2019objects} with association based on IoU matching~\cite{bewley2016simple,yang2023hard} or Re-ID features~\cite{zhang2021fairmot}.

Recent benchmarks such as DanceTrack~\cite{sun2022dancetrack} and SportsMOT~\cite{cui2023sportsmot} emphasize crowded scenes, frequent occlusions, complex motion, and appearance-homogeneous targets.
These settings expose limitations of frame-wise processing with heuristic association and motivate models that learn temporal association within the network.
Transformer-based trackers such as TrackFormer~\cite{meinhardt2022trackformer} and MOTR~\cite{zeng2022motr} propagate track queries across frames and update them through attention to frame features.
Later methods improve query-based tracking with detector priors in MOTRv2~\cite{zhang2023motrv2}, long-term memory in MeMOTR~\cite{gao2023memotr}, and efficient long-sequence modeling in SambaMOTR~\cite{segu2025samba} based on selective state-space models~\cite{gu2024mamba}.

Overall, MOT has evolved from post-hoc association on per-frame detections to end-to-end frameworks that maintain internal track states across frames.
QPID leverages these propagated track states to estimate association instability for clip selection, rather than modifying the tracker architecture itself.

\subsection{Active Learning}
\label{sec_related_al}

Active Learning aims to reduce annotation cost by selecting informative samples from an unlabeled pool.
In computer vision, active learning has been studied for image classification~\cite{beluch2018power,ducoffe2018adversarial,wang2016cost}, object detection~\cite{brust2018active,schmidt2020advanced,yang2024plug}, semantic segmentation~\cite{casanova2020reinforced,mittal2023best}, and vision-language settings~\cite{bang2024active,safaei2025active}.
Common acquisition strategies include uncertainty-based sampling~\cite{wang2016cost,ducoffe2018adversarial}, representative selection such as core-set methods~\cite{sener2017active}, and hybrid strategies that combine uncertainty and diversity~\cite{ash2019deep,citovsky2021batch}.
Recent coverage-based methods further formalize batch selection through generalized coverage and uncertainty coverage~\cite{bae2024generalized,bae2025uncertainty}.
QPID is related to this coverage-based perspective, but adapts it to clip-level MOT by selecting a visually representative subset from high-instability clips using track-level visual prototypes.

\noindent\textbf{Perturbation- and consistency-based AL.}
Several methods estimate informativeness by measuring prediction changes under input perturbations or augmentations.
In object detection, Localization Stability~\cite{kao2018localization} measures how predicted object locations vary when input images are corrupted by noise, while CALD~\cite{yu2022consistency} compares predictions between original and augmented images using box and class-score consistency.
These methods probe detector robustness to input-space transformations, whereas QPID perturbs internal track states of query-propagation trackers to estimate association instability.

\noindent\textbf{Active learning for videos.}
Extending active learning to videos is challenging because samples are temporally correlated and annotations are structured across frames.
Several video AL methods reduce annotation cost by selecting informative videos or frames for action detection, video classification, object detection, and video segmentation~\cite{rana2022all,zolfaghari2019temporal,goswami2023active,rana2023hybrid,singh2024semi,wu2024correlation}.
However, frame-level acquisition is not fully aligned with end-to-end MOT trackers, whose training samples are multi-frame clips with temporally consistent identities.
A single frame may not reveal whether the tracker can maintain identity through temporal interactions such as crossing, occlusion, and reappearance.

\noindent\textbf{Active learning for MOT.}
Active learning for MOT has been less explored.
HD-AMOT~\cite{li2023heterogeneous} formulates informative frame selection as a Markov Decision Process using heterogeneous geometric and semantic cues.
It performs frame-level sampling, whereas QPID selects multi-frame clips aligned with the training structure of end-to-end MOT trackers.
SPAM~\cite{cetintas2024spamming} improves tracking annotation efficiency by combining pseudo labels, human correction, and a graph-based labeling model.
Unlike SPAM, which is designed as a video label engine for producing tracking annotations, QPID focuses on selecting which clips should be annotated to train end-to-end trackers.

CUTAL~\cite{inoue2026clip} is the most directly related work, formulating clip-level active learning for end-to-end MOT with output-level temporal uncertainty and temporal diversity sampling.
QPID differs from CUTAL by probing internal track states through two-sided perturbations to estimate association instability beyond final-prediction variation.
It further uses track-level visual coverage to select informative and non-redundant clips.
\section{Problem Definition}
\label{sec_prob_def}

We follow the clip-level active learning setting introduced by CUTAL for end-to-end MOT.
Given a set of training videos \(\mathcal{V}\), a clip pool \(\mathcal{C}\) is constructed by sampling fixed-length clips from each video.
Each clip \(c\in\mathcal{C}\) is associated with a source video \(\nu(c)\in\mathcal{V}\), starts at frame \(t_c\), and consists of \(T\) sampled frames with an intra-clip interval \(\Delta\):
\begin{equation}
\mathcal{K}(c)
=
\{t_c+j\Delta\mid j=0,\dots,T-1\}.
\end{equation}
Here, \(T\) is chosen to match the standard training clip length of the target end-to-end tracker.
Annotating a clip requires bounding boxes and temporally consistent identity labels for all sampled frames in \(\mathcal{K}(c)\), so the annotation cost is proportional to \(T\).

Active learning proceeds over rounds.
At round \(r\), the labeled set is \(\mathcal{L}_r\subset\mathcal{C}\), and the unlabeled pool is \(\mathcal{U}_r=\mathcal{C}\setminus\mathcal{L}_r\).
The process starts from a small random seed set \(\mathcal{L}_0\).
At each round, the tracker is trained on \(\mathcal{L}_r\), acquisition scores are computed for clips in \(\mathcal{U}_r\), and a subset \(\mathcal{S}_r\subset\mathcal{U}_r\) is selected for annotation.

Given a frame-level annotation budget \(b_r\), the corresponding clip budget is \(B_r=\lfloor b_r/T\rfloor\).
The selected batch satisfies \(\mathcal{S}_r\subset\mathcal{U}_r\) and \(|\mathcal{S}_r|=B_r\).
After annotation, the labeled set and unlabeled pool are updated as \(\mathcal{L}_{r+1}=\mathcal{L}_r\cup\mathcal{S}_r\) and \(\mathcal{U}_{r+1}=\mathcal{U}_r\setminus\mathcal{S}_r\).

Following CUTAL, clips that temporally overlap with already labeled clips from the same video are excluded from selection.
We also enforce temporal disjointness among clips selected within the same round to avoid duplicate annotation.

\section{Method}
\label{sec_method}

We propose \textbf{QPID}, a clip acquisition method for query-propagation MOT under the clip-level active learning setting.
While CUTAL estimates clip informativeness from output-level temporal uncertainty, QPID targets association instability in propagated track states.
QPID applies two-sided perturbations to internal track states and measures perturbation-induced prediction differences from a clean reference branch.
This enables QPID to prioritize clips where small changes to propagated track states produce large prediction differences, even when final predictions appear temporally smooth.

\begin{figure}[t]
 \centering
 \includegraphics[width=1.0\linewidth]{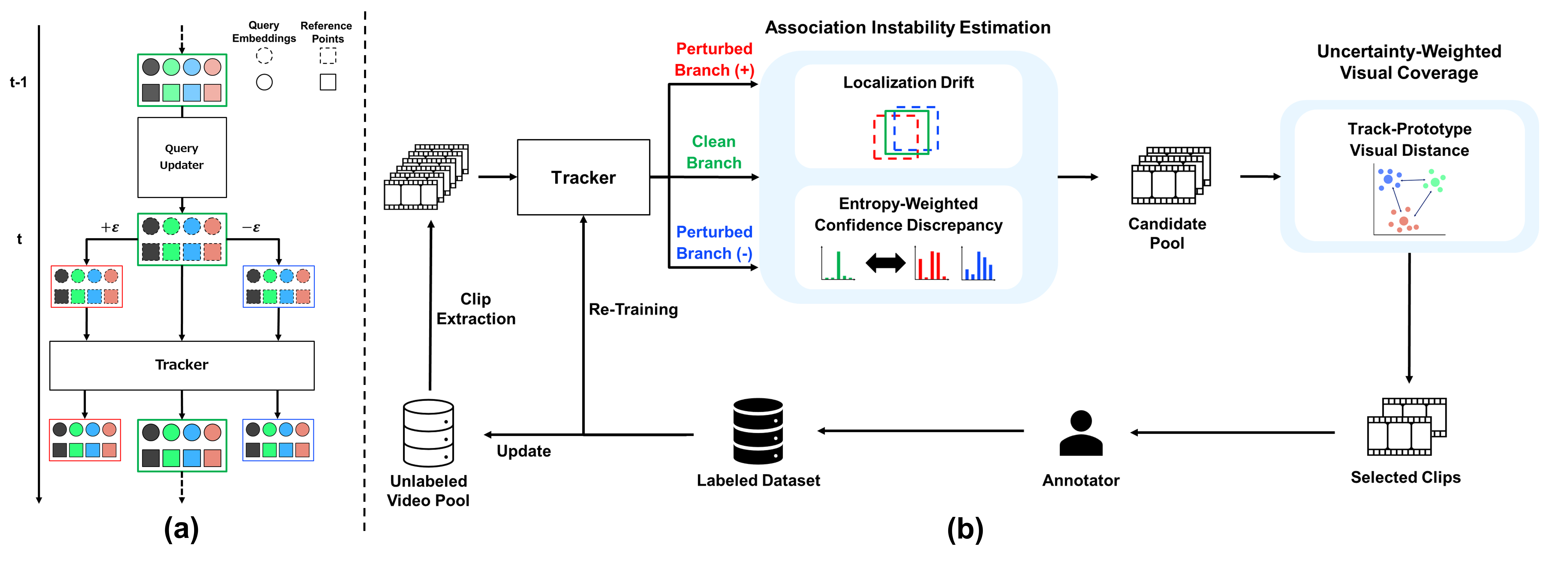}
    \caption{
    \textbf{Overview of QPID.}
    \textbf{(a) Two-sided track-state perturbation.}
    QPID applies positive and negative perturbations along the same sampled direction to internal track states and compares the resulting predictions with a clean reference branch.
    Perturbed states are not propagated to subsequent frames, so the score reflects local perturbation-induced prediction differences.
    \textbf{(b) Clip acquisition pipeline.}
    QPID aggregates Localization Drift and Entropy-Weighted Confidence Discrepancy into a clip-level association-instability score, forms a high-instability candidate pool, and selects a representative annotation batch by Uncertainty-Weighted Visual Coverage with track-level visual prototypes.
    }
 \label{fig:fig2}
\end{figure}

As shown in Fig.~\ref{fig:fig2}(b), QPID consists of two stages.
First, \textit{Two-Sided Track-State Perturbation} computes a clip-level association-instability score \(U(c)\) from \textit{Localization Drift} and \textit{Entropy-Weighted Confidence Discrepancy}.
Second, QPID constructs a high-instability candidate pool and selects a representative annotation batch by \textit{Uncertainty-Weighted Visual Coverage} using track-level visual prototypes.

\subsection{Two-Sided Track-State Perturbation}
\label{sec_method_noise}

In query-propagation MOT, propagated track states interact with current visual features at each frame and are updated to produce track predictions.
If association is stable, small changes to the track state should produce consistent predictions for the same propagated track.
In contrast, when appearance-similar targets approach or cross each other, small changes to the track state can lead to localization drift or confidence changes.

QPID measures these changes by comparing a clean reference branch with positive and negative perturbed branches at each frame \(t\).
The clean branch uses the track state propagated from frame \(t-1\).
For track instance \(i\), we denote the clean track query embedding by \(q^0_{i,t}\in\mathbb{R}^d\) and the normalized reference point by:
\begin{equation}
\tilde r^0_{i,t}
=
(\tilde c^0_{x,i,t},\tilde c^0_{y,i,t},\tilde w^0_{i,t},\tilde h^0_{i,t})
\in[0,1]^4.
\end{equation}
Here, \((\tilde c_x,\tilde c_y)\) denotes the reference-point center and \((\tilde w,\tilde h)\) denotes the reference-point width and height~\cite{liu2022dab}.
We refer to the pair of the track query embedding and the reference point as an internal track state.
The track query embedding is a propagated track-specific representation, while the reference point provides a spatial anchor for localization.
By perturbing both components, QPID probes whether the propagated track state remains stable in ambiguous association cases.

We use two-sided perturbations by applying the same sampled perturbation direction with opposite signs to probe the local response around the clean state.
From the clean state \((q^0_{i,t},\tilde r^0_{i,t})\), QPID constructs a positive perturbed state \((q^+_{i,t},\tilde r^+_{i,t})\) and a negative perturbed state \((q^-_{i,t},\tilde r^-_{i,t})\).
The clean, positive, and negative branches are initialized from the same clean track instance \(i\) at frame \(t\), so QPID compares predictions with the same track index across branches without additional matching (Fig.~\ref{fig:fig2}(a)).
After scoring frame \(t\), the perturbed branches are discarded, and only the clean state is propagated to frame \(t+1\).
Thus, perturbations do not accumulate over time, and QPID measures local perturbation-induced prediction differences at each sampled frame.

\noindent\textbf{Query-embedding perturbation.}
For each clip and track instance, QPID samples one perturbation direction and reuses the same direction across the sampled frames.
To control the perturbation magnitude across tracks and clips, we sample \(z_{q,i}\sim\mathcal{N}(0,\mathbf{I}_d)\) and project it onto the \(\ell_2\) sphere with radius \(\alpha_q\):
\begin{equation}
\varepsilon_{q,i}
=
\alpha_q
\frac{z_{q,i}}{\lVert z_{q,i}\rVert_2}.
\end{equation}
The positive and negative perturbed queries are:
\begin{equation}
q^+_{i,t}
=
q^0_{i,t}
+
\varepsilon_{q,i},
\qquad
q^-_{i,t}
=
q^0_{i,t}
-
\varepsilon_{q,i}.
\end{equation}
Here, \(\mathbf{I}_d\) is the \(d\times d\) identity matrix.

\noindent\textbf{Scale-calibrated reference-point perturbation.}
Adding a fixed offset to the normalized reference point can introduce scale bias.
The same absolute displacement corresponds to a larger relative change for small objects than for large objects.
To mitigate this bias, QPID scales the reference-point perturbation by the width and height of the clean propagated reference point.
We sample \(z_{\tilde r,i}\sim\mathcal{N}(0,\mathbf{I}_4)\) and define:
\begin{equation}
\varepsilon_{\tilde r,i}
=
\alpha_{\tilde r}
\frac{z_{\tilde r,i}}{\lVert z_{\tilde r,i}\rVert_2}.
\end{equation}
The perturbed reference points are:
\begin{equation}
\tilde r^+_{i,t}
=
\Pi_{[0,1]^4}
\left(
\tilde r^0_{i,t}
+
(\tilde w^0_{i,t},\tilde h^0_{i,t},\tilde w^0_{i,t},\tilde h^0_{i,t})
\odot
\varepsilon_{\tilde r,i}
\right),
\end{equation}
\begin{equation}
\tilde r^-_{i,t}
=
\Pi_{[0,1]^4}
\left(
\tilde r^0_{i,t}
-
(\tilde w^0_{i,t},\tilde h^0_{i,t},\tilde w^0_{i,t},\tilde h^0_{i,t})
\odot
\varepsilon_{\tilde r,i}
\right),
\end{equation}
where \(\Pi_{[0,1]^4}\) denotes element-wise clipping to the range \([0,1]^4\), and \(\odot\) denotes element-wise multiplication.

\subsection{Association-Aware Instability Metrics}
\label{sec_method_metrics}

QPID quantifies perturbation-induced prediction differences using the tracker outputs.
For branch \(k\in\{0,+,-\}\), let \(b^k_{i,t}=(c^k_{x,i,t},c^k_{y,i,t},w^k_{i,t},h^k_{i,t})\) be the predicted bounding box of track instance \(i\) at frame \(t\), and let \(s^k_{i,t}\) be its confidence score.
The metrics are computed only for active tracks in the clean branch:
\begin{equation}
\mathcal{I}_t
=
\{i\mid s^0_{i,t}>\gamma_{\mathrm{score}}\}.
\end{equation}
Here, \(\gamma_{\mathrm{score}}\) is the confidence threshold for selecting active clean-branch tracks, and we use \(\gamma_{\mathrm{score}}=0.5\) in all experiments.

\noindent\textbf{Localization Drift.}
Localization Drift measures how much the predicted box changes under two-sided perturbations.
It is defined as the average IoU drop between the clean prediction and the perturbed predictions:
\begin{equation}
D_{i,t}
=
\frac{1}{2}
\sum_{k\in\{+,-\}}
\left(
1
-
\mathrm{IoU}
\left(
b^0_{i,t},
b^k_{i,t}
\right)
\right).
\end{equation}
A large \(D_{i,t}\) indicates that a small perturbation to the internal track state causes a large localization change.

\noindent\textbf{Entropy-Weighted Confidence Discrepancy.}
Perturbation-induced instability can also appear as changes in confidence scores.
However, raw confidence differences alone do not distinguish informative uncertainty from small fluctuations in already confident predictions.
QPID therefore weights the confidence discrepancy by the binary entropy of the clean-branch confidence:
\begin{equation}
\mathcal{H}(s)
=
-
s\log s
-
(1-s)\log(1-s).
\end{equation}
The Entropy-Weighted Confidence Discrepancy is defined as:
\begin{equation}
E_{i,t}
=
\mathcal{H}(s^0_{i,t})
\cdot
\frac{1}{2}
\sum_{k\in\{+,-\}}
\left|
s^0_{i,t}
-
s^k_{i,t}
\right|.
\end{equation}
This weighting emphasizes perturbation-induced confidence changes when the clean prediction is uncertain and suppresses fluctuations for highly confident predictions.

\subsection{Score Aggregation and Normalization}
\label{sec_method_agg}

Since annotation is performed at the clip level, instance-level metrics must be aggregated into a clip-level score.
QPID emphasizes hard cases within a clip by taking the maximum over active track instances at each sampled frame and averaging these values over the clip.
For metric \(X\in\{D,E\}\), let \(X_{i,t}\) be its value for track instance \(i\) at frame \(t\).
The aggregated score for clip \(c\) is:
\begin{equation}
\bar{X}(c)
=
\frac{1}{|\mathcal{K}(c)|}
\sum_{t\in\mathcal{K}(c)}
\max_{i\in\mathcal{I}_t}
X_{i,t}.
\end{equation}
If \(\mathcal{I}_t=\emptyset\), we set \(\max_{i\in\mathcal{I}_t}X_{i,t}=0\).

Following CUTAL~\cite{inoue2026clip}, QPID normalizes each component using statistics over the unlabeled pool \(\mathcal{U}_r\) at each round.
This normalization makes the two components comparable, since localization drift and confidence discrepancy have different ranges and distributions.
For metric \(X\), let \(\mu_X\) and \(\sigma_X\) be the mean and standard deviation of \(\{\bar{X}(c)\}_{c\in\mathcal{U}_r}\).
We define a clipped normalization function:
\begin{equation}
\phi_X(x)
=
\max
\left(
0,
\min
\left(
1,
\frac{x-(\mu_X-\kappa\sigma_X)}
{2\kappa\sigma_X}
\right)
\right).
\end{equation}
Here, \(\kappa\) controls the clipping range.

The final clip-level association-instability score is:
\begin{equation}
U(c)
=
\phi_D(\bar{D}(c))
\cdot
\phi_E(\bar{E}(c)).
\end{equation}
We use multiplicative aggregation to prioritize clips where perturbations jointly affect localization and confidence, and analyze an additive alternative in the ablation study.

\subsection{Track-Prototype Visual Distance}
\label{sec_method_visual_distance}

The association-instability score \(U(c)\) ranks clips by estimated association instability, but batch selection also needs to avoid visual redundancy.
Since a MOT clip contains multiple targets, a single global clip feature can obscure which tracked objects are redundant.
QPID therefore represents each clip by a set of track-level visual prototypes and compares clips with a symmetric weighted Chamfer distance.

We use the final-layer decoder output embedding \(e^0_{i,t}\in\mathbb{R}^d\) of track instance \(i\) at frame \(t\) in the clean branch.
Let \(\mathrm{norm}(x)=x/\|x\|_2\) denote \(\ell_2\) normalization.
For clip \(c\), the active frames of track instance \(i\) are:
\begin{equation}
\mathcal{K}_i(c)
=
\{t\in\mathcal{K}(c)\mid s^0_{i,t}>\gamma_{\mathrm{score}}\}.
\end{equation}
If \(\mathcal{K}_i(c)\neq\emptyset\), the track prototype is:
\begin{equation}
\bar{e}_i(c)
=
\mathrm{norm}
\left(
\frac{1}{|\mathcal{K}_i(c)|}
\sum_{t\in\mathcal{K}_i(c)}
\mathrm{norm}(e^0_{i,t})
\right).
\end{equation}

To reflect prototype reliability, QPID weights each prototype by the average clean-branch confidence over its active frames:
\begin{equation}
\omega_i(c)
=
\frac{1}{|\mathcal{K}_i(c)|}
\sum_{t\in\mathcal{K}_i(c)}
s^0_{i,t}.
\end{equation}
After normalizing weights within each clip, we obtain the weighted prototype set
\(\mathcal{P}(c)=\{(\bar e_u(c),\bar\omega_u(c))\}_{u=1}^{N_c}\).
Here, \(N_c\) denotes the number of valid prototypes, and the normalized weights satisfy
\(\sum_{u=1}^{N_c}\bar\omega_u(c)=1\).

For two clips \(c_m\) and \(c_n\), the Track-Prototype Visual Distance is defined as:
\begin{equation}
\begin{aligned}
\mathrm{Dist}_{\mathrm{vis}}(c_m,c_n)
=
&
\sum_{u=1}^{N_{c_m}}
\bar{\omega}_u(c_m)
\min_{v\in\{1,\dots,N_{c_n}\}}
d_{\cos}
\left(
\bar{e}_u(c_m),
\bar{e}_v(c_n)
\right)
\\
&
+
\sum_{v=1}^{N_{c_n}}
\bar{\omega}_v(c_n)
\min_{u\in\{1,\dots,N_{c_m}\}}
d_{\cos}
\left(
\bar{e}_v(c_n),
\bar{e}_u(c_m)
\right),
\end{aligned}
\end{equation}
where \(d_{\cos}(a,b)=1-a^\top b\) denotes cosine distance for normalized feature vectors.
This distance measures visual redundancy at the tracked-object level, making it more suitable for multi-target MOT clips than a single global clip representation.

\subsection{Uncertainty-Weighted Visual Coverage}
\label{sec_method_coverage}

Using the Track-Prototype Visual Distance, QPID selects a representative batch from clips with high association instability.
The selection proceeds in two steps.
First, QPID restricts the search space to high-instability clips.
At round \(r\), given the clip budget \(B_r\) and the candidate expansion ratio \(\rho>1\), 
we form the high-instability candidate pool 
\(\mathcal{C}^{\mathrm{cand}}_r\) by selecting up to 
\(\rho B_r\) clips in \(\mathcal{U}_r\) with the largest association-instability scores \(U(c)\).

Second, QPID greedily selects clips to improve visual coverage of the high-instability candidate pool.
Let \(\mathcal{S}\) denote the clips already selected in the current round, initialized as \(\emptyset\).
The current reference set is defined as:
\begin{equation}
\mathcal{A}(\mathcal{S})
=
\mathcal{L}_r
\cup
\mathcal{S}.
\end{equation}
For each clip \(c\in\mathcal{C}^{\mathrm{cand}}_r\), its distance to the current reference set is:
\begin{equation}
d_{\mathcal{A}}(c)
=
\min_{a\in\mathcal{A}(\mathcal{S})}
\mathrm{Dist}_{\mathrm{vis}}(c,a).
\end{equation}
A larger \(d_{\mathcal{A}}(c)\) means that \(c\) is less covered by the labeled or already selected clips.

If a clip \(c'\) is selected, the nearest distance of each clip \(c\in\mathcal{C}^{\mathrm{cand}}_r\) is updated as:
\begin{equation}
d_{\mathcal{A}\cup\{c'\}}(c)
=
\min
\left\{
d_{\mathcal{A}}(c),
\mathrm{Dist}_{\mathrm{vis}}(c,c')
\right\}.
\end{equation}
QPID measures the benefit of selecting \(c'\) by the uncertainty-weighted reduction of these distances:
\begin{equation}
G(c')
=
\sum_{c\in\mathcal{C}^{\mathrm{cand}}_r}
U(c)
\left[
d_{\mathcal{A}}(c)
-
d_{\mathcal{A}\cup\{c'\}}(c)
\right].
\end{equation}
Thus, a clip is favored when it improves coverage for many high-instability clips.

At each greedy step, QPID selects:
\begin{equation}
c^\ast
=
\operatorname*{argmax}_{c'\in\mathcal{C}^{\mathrm{cand}}_r\setminus\mathcal{S}}
G(c').
\end{equation}
Clips that temporally overlap with labeled clips or already selected clips from the same video are excluded from this maximization.
The selected clip \(c^\ast\) is added to \(\mathcal{S}\), and the procedure is repeated until \(|\mathcal{S}|=B_r\).
The final selected set is used as the annotation batch \(\mathcal{S}_r\).
\section{Experiments}
\label{sec:experiments}
\suppressfloats[t]
\vspace{-1mm}
We evaluate QPID in a clip-level active learning setting for end-to-end MOT.
Additional round-wise results, ablations, statistical analyses, qualitative examples, and implementation details are provided in the Supplementary Material.

\begin{figure}[t]
    \centering

    \begin{adjustbox}{width=0.9\linewidth,center}
        \begin{minipage}[t]{0.32\linewidth}
            \centering
            \includegraphics[width=\linewidth]{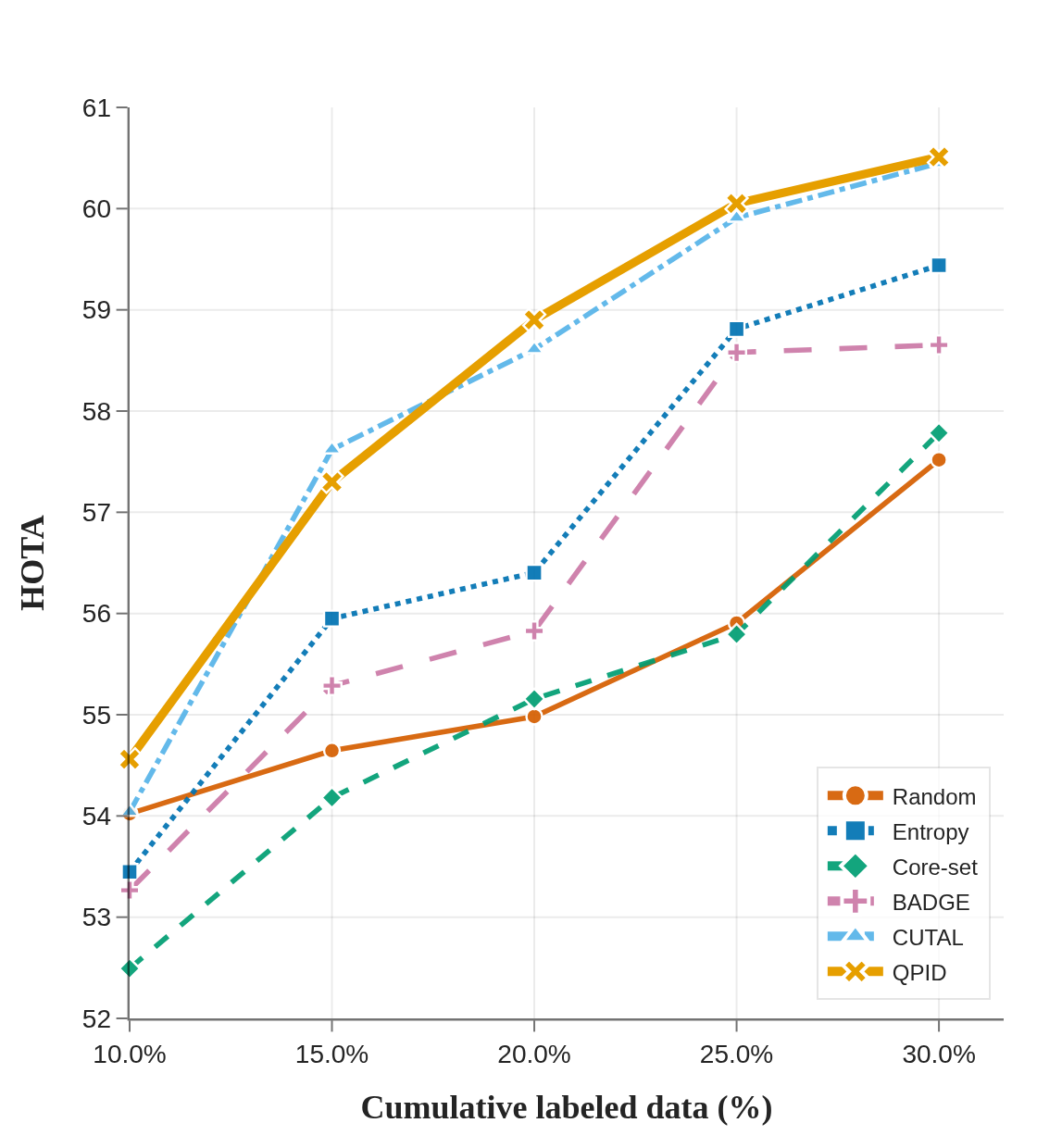}
        \end{minipage}
        \hfill
        \begin{minipage}[t]{0.32\linewidth}
            \centering
            \includegraphics[width=\linewidth]{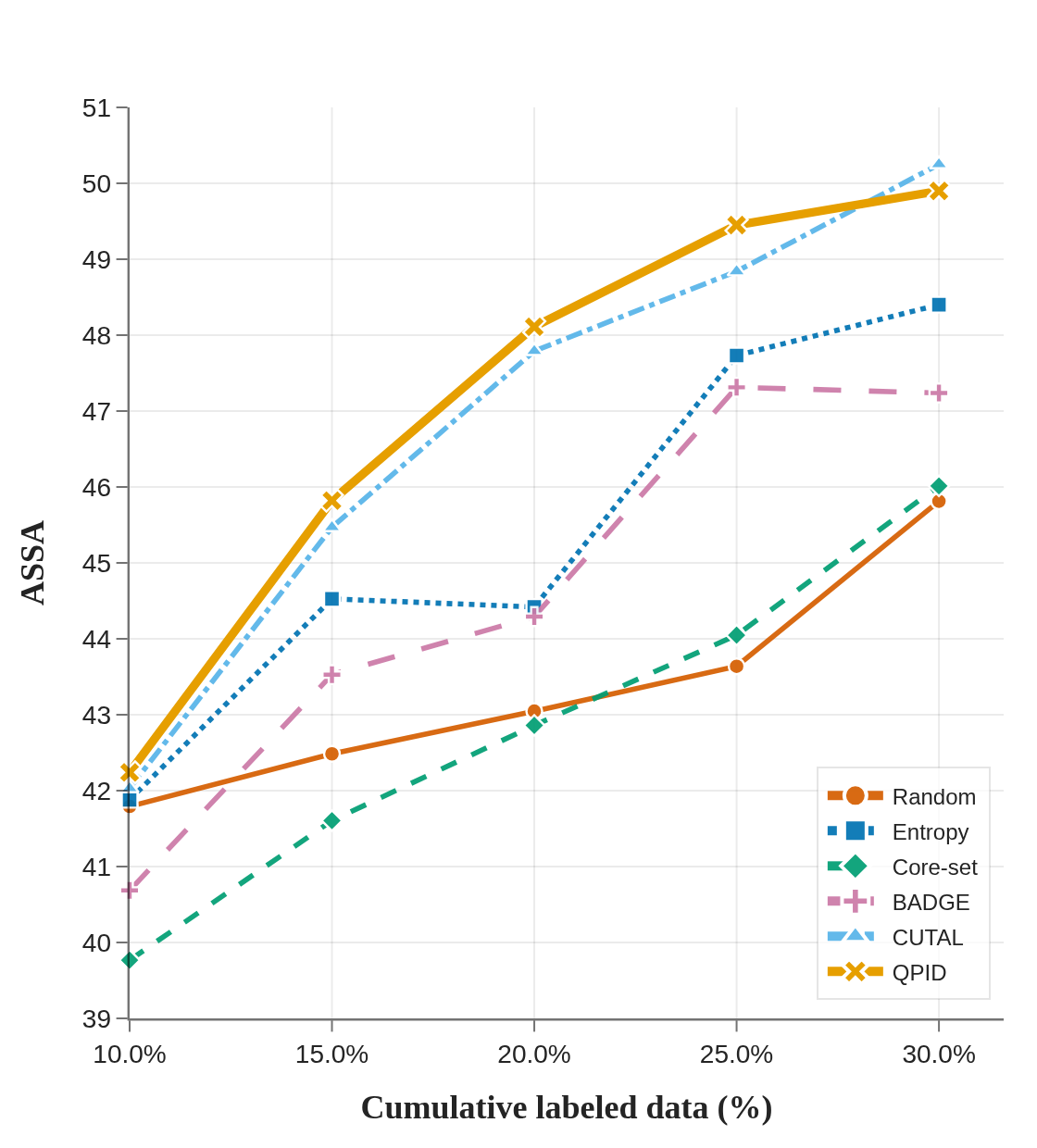}
        \end{minipage}
        \hfill
        \begin{minipage}[t]{0.32\linewidth}
            \centering
            \includegraphics[width=\linewidth]{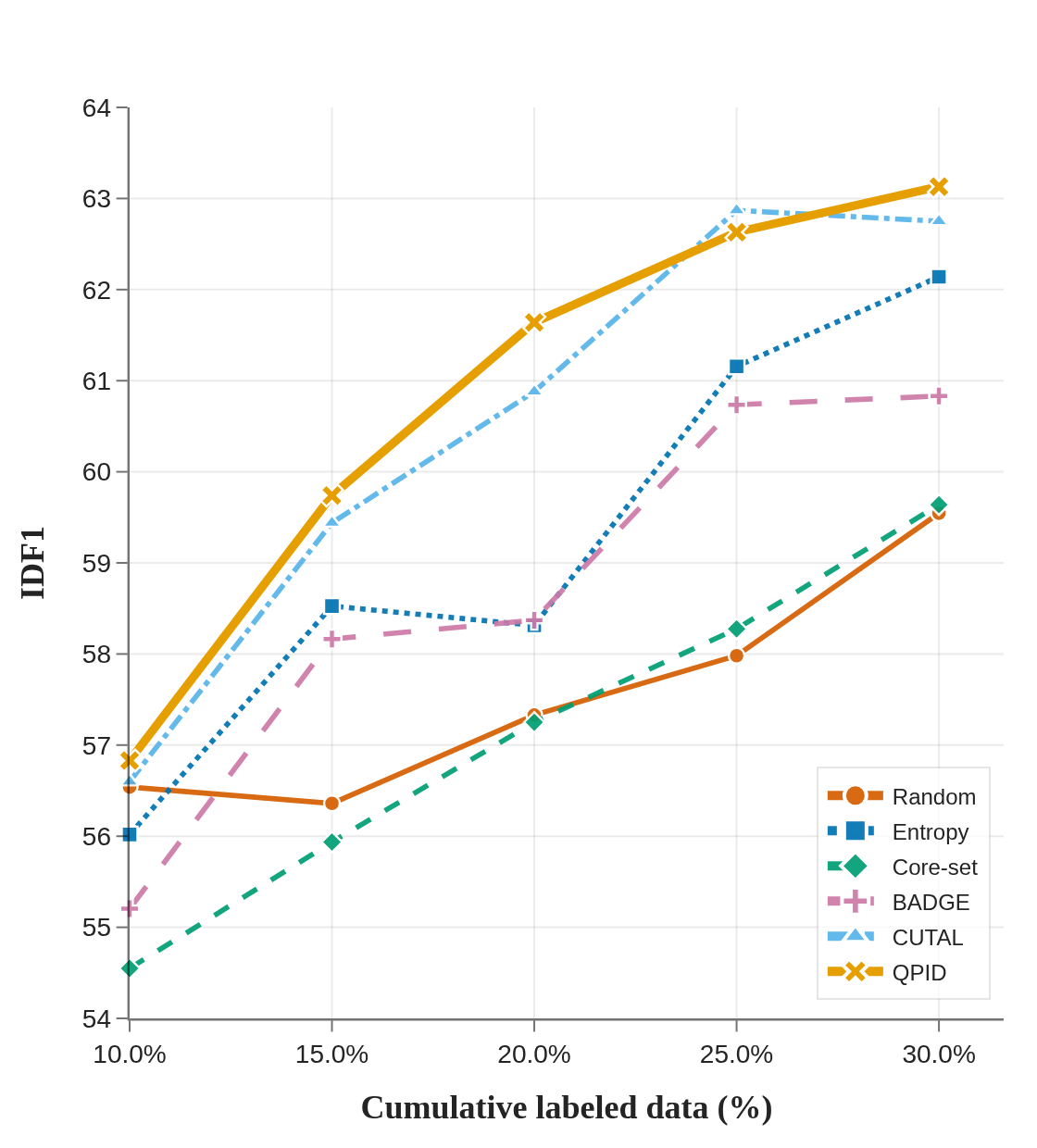}
        \end{minipage}
    \end{adjustbox}

    \vspace{-1.2mm}
    \begin{adjustbox}{width=0.9\linewidth,center}
        \begin{minipage}[t]{0.32\linewidth}
            \centering
            \includegraphics[width=\linewidth]{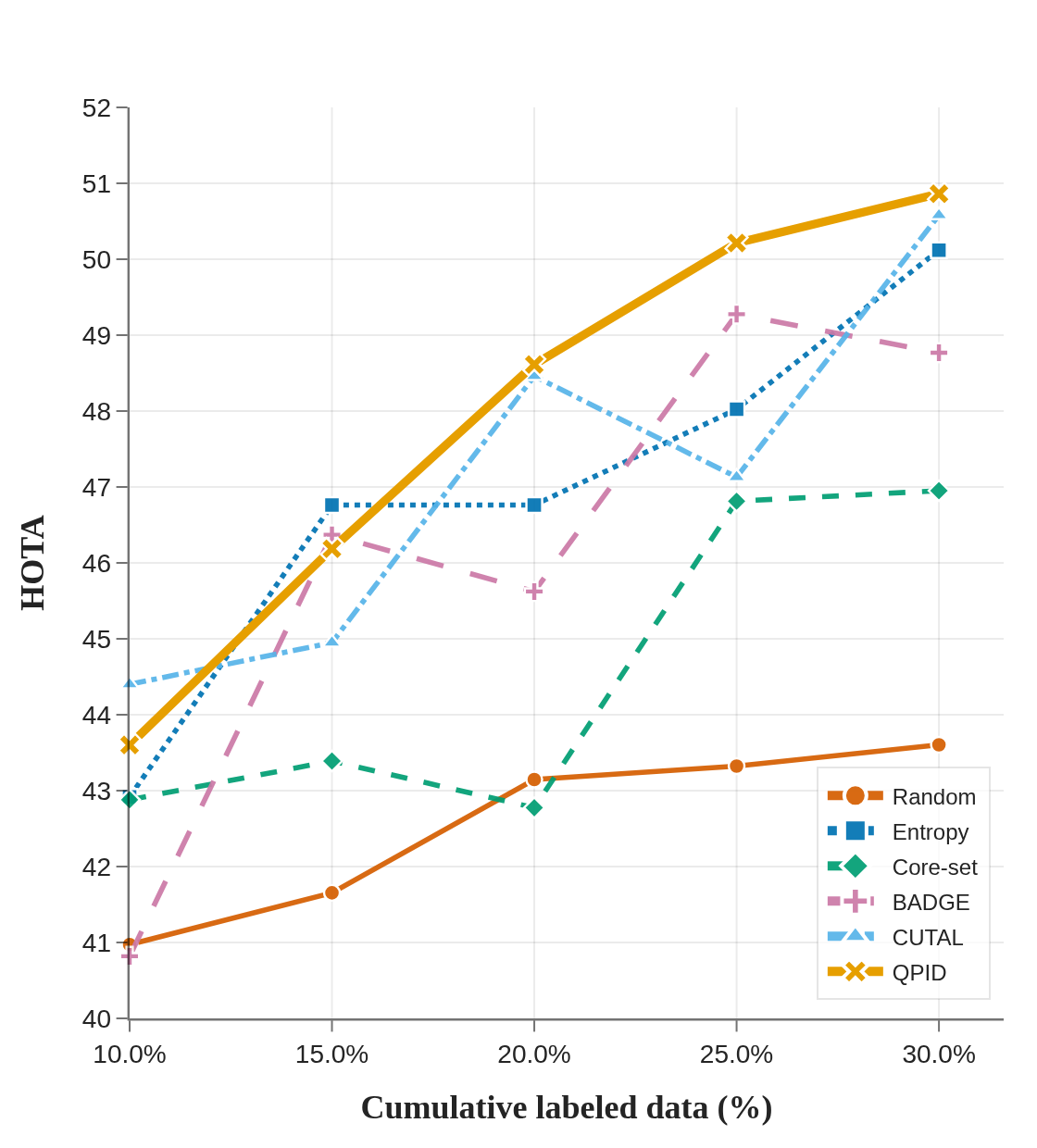}
        \end{minipage}
        \hfill
        \begin{minipage}[t]{0.32\linewidth}
            \centering
            \includegraphics[width=\linewidth]{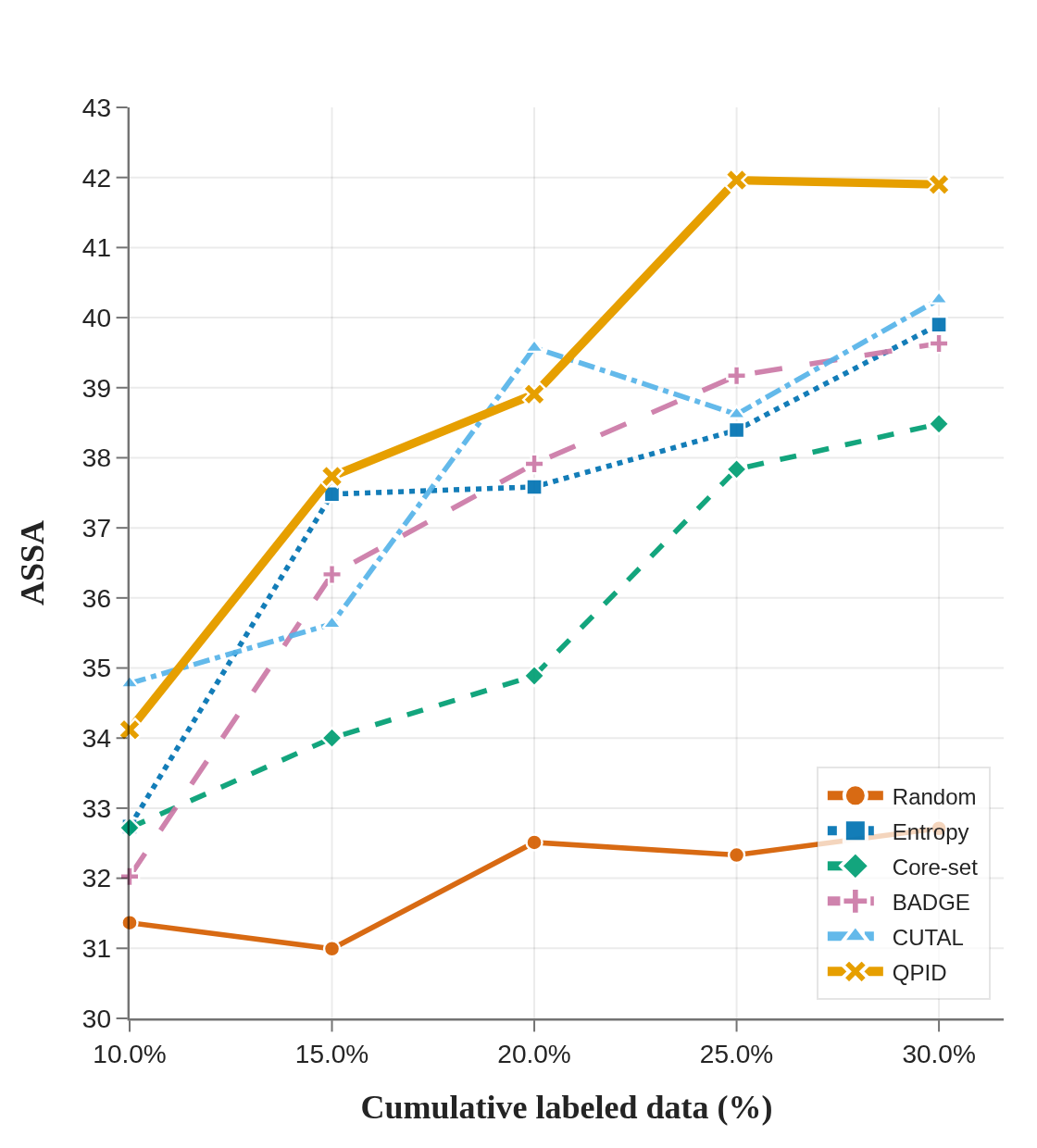}
        \end{minipage}
        \hfill
        \begin{minipage}[t]{0.32\linewidth}
            \centering
            \includegraphics[width=\linewidth]{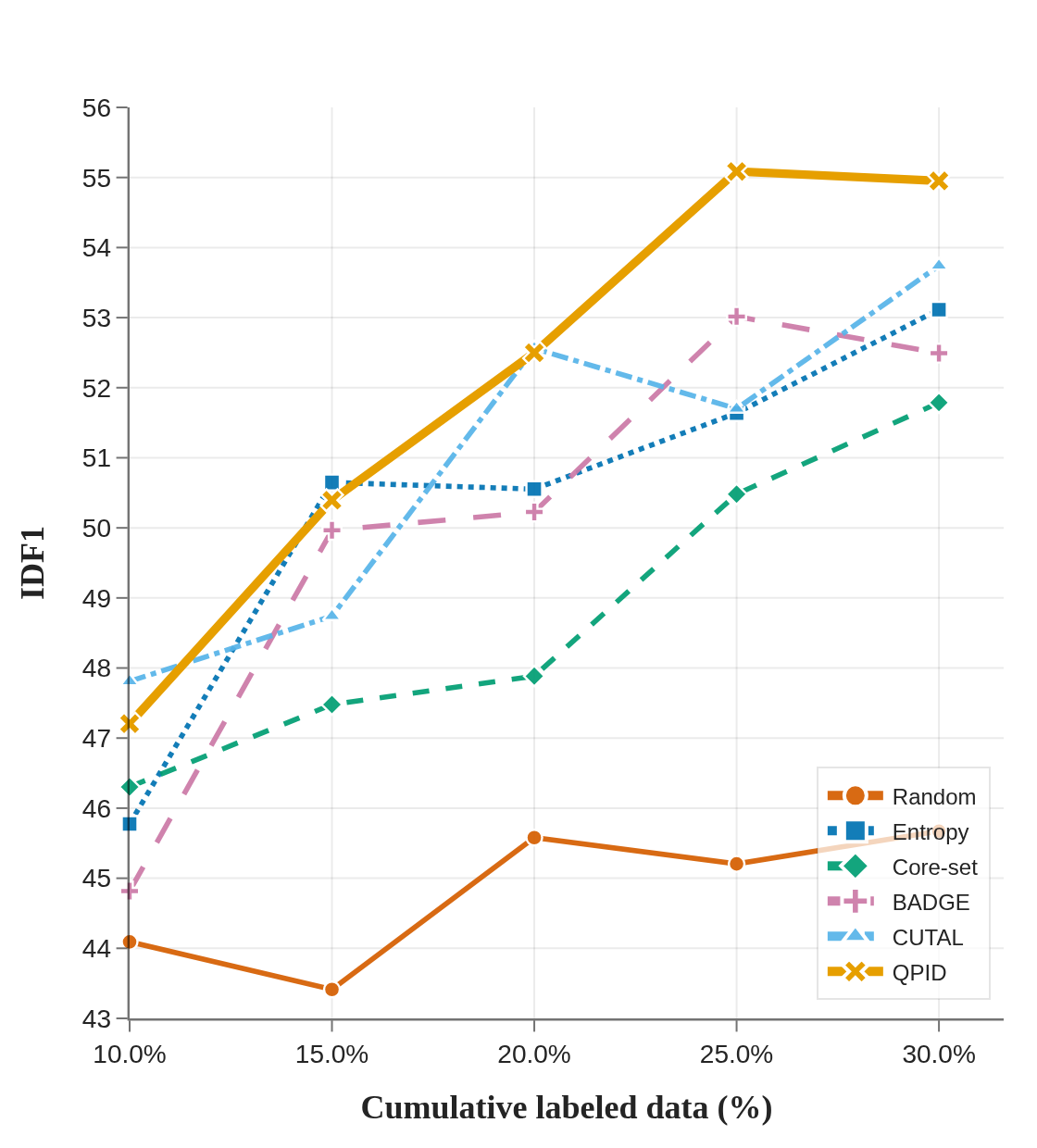}
        \end{minipage}
    \end{adjustbox}
    \vspace{-3mm}
    \caption{\textbf{Round-wise performance on DanceTrack under the 5\%+5\% schedule.} The upper and lower rows show results using MeMOTR and SambaMOTR, respectively. From left to right, we report HOTA, AssA, and IDF1 of QPID and baseline methods.}
    \label{fig:memotr_samba_dancetrack_55}
\end{figure}

\begin{figure}[t]
    \centering

    \begin{adjustbox}{width=0.9\linewidth,center}
        \begin{minipage}[t]{0.32\linewidth}
            \centering
            \includegraphics[width=\linewidth]{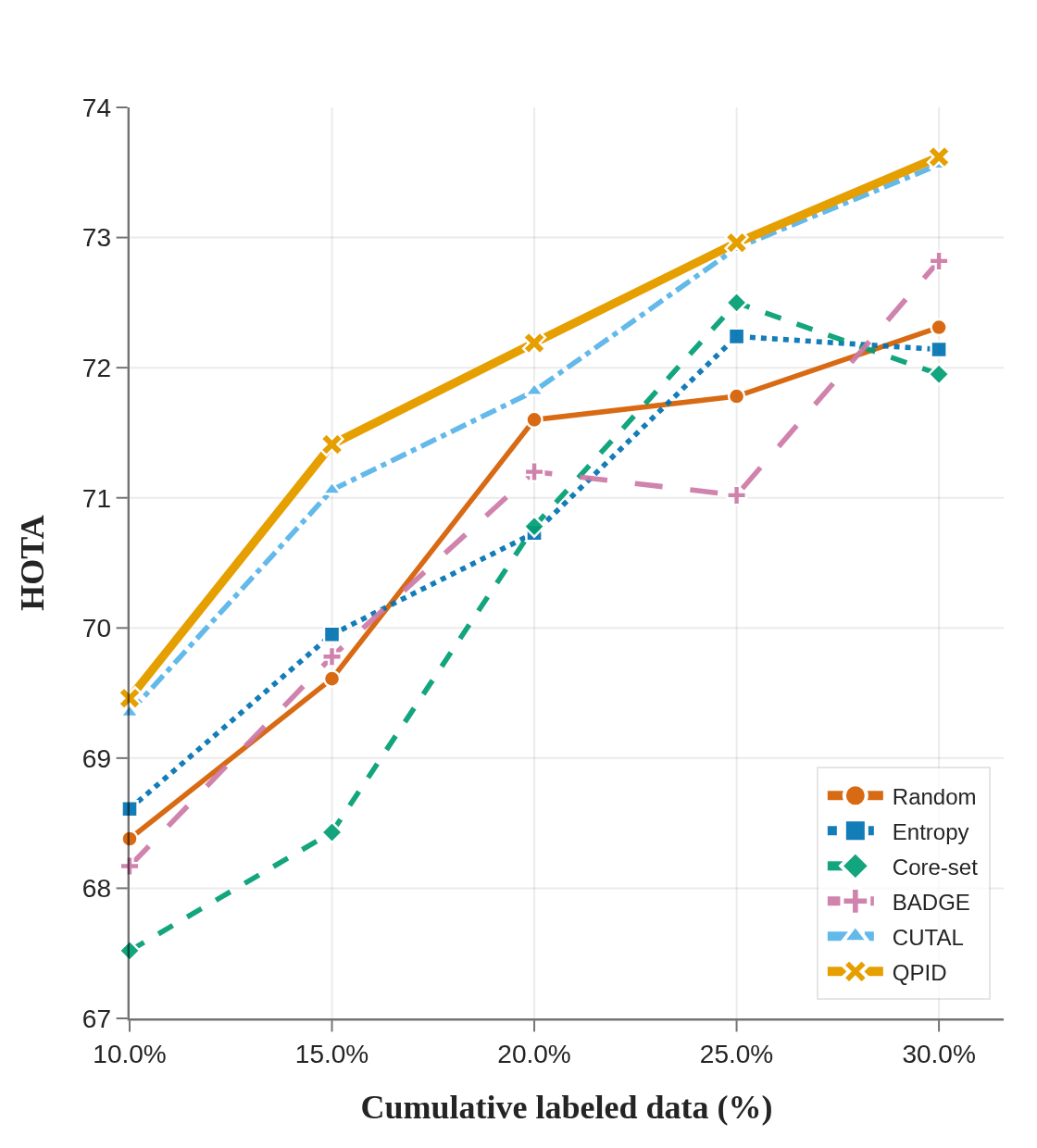}
        \end{minipage}
        \hfill
        \begin{minipage}[t]{0.32\linewidth}
            \centering
            \includegraphics[width=\linewidth]{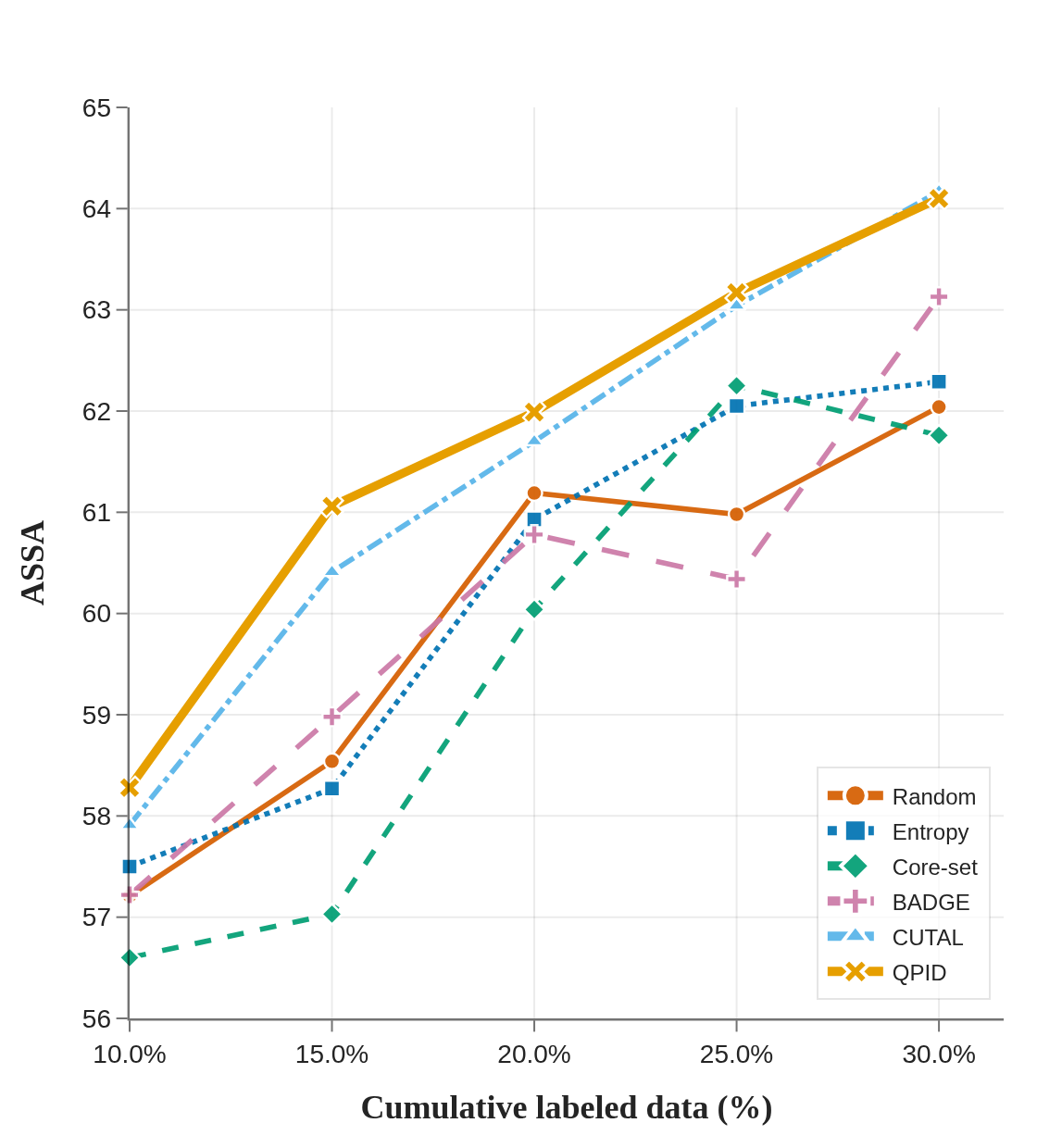}
        \end{minipage}
        \hfill
        \begin{minipage}[t]{0.32\linewidth}
            \centering
            \includegraphics[width=\linewidth]{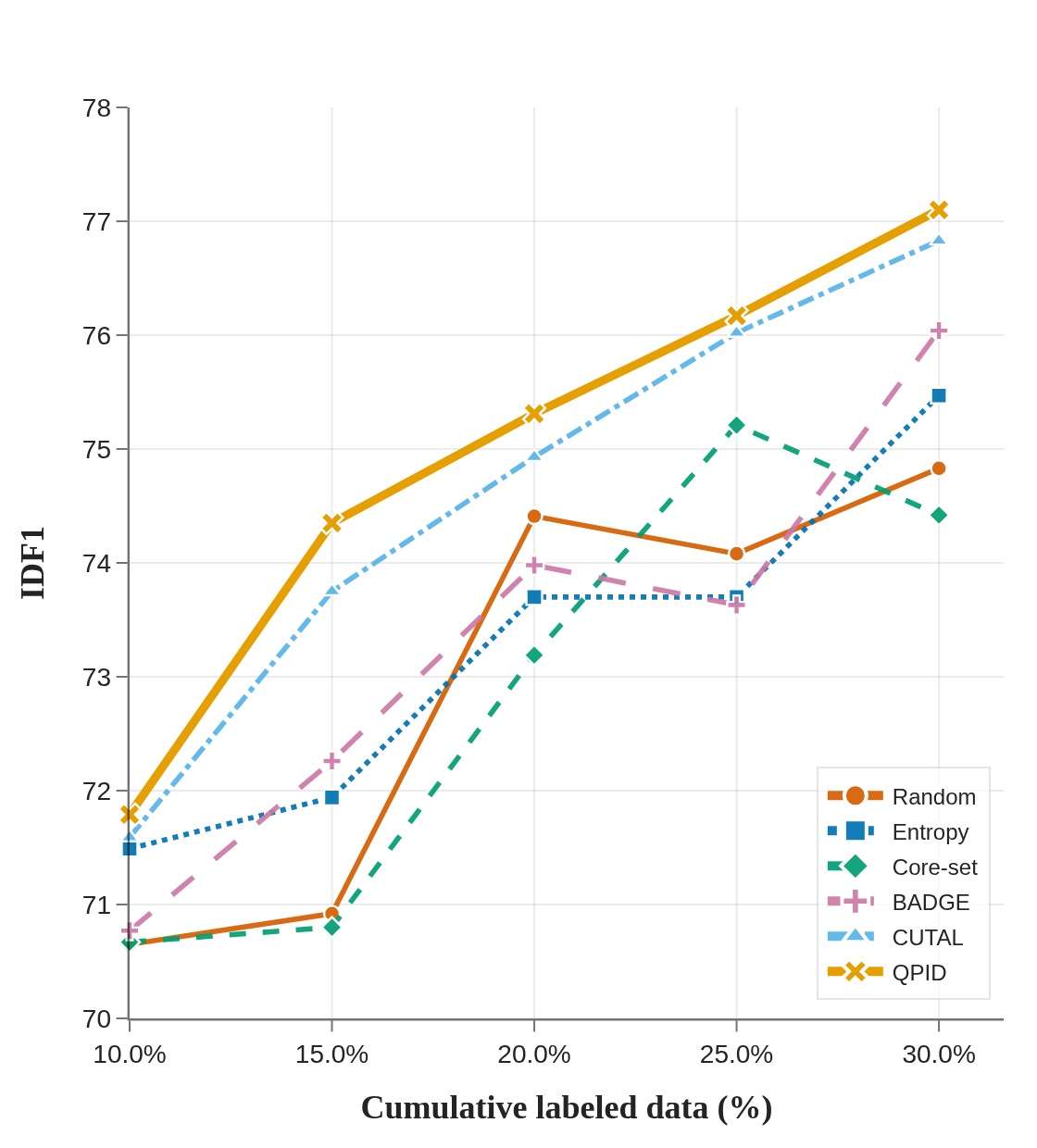}
        \end{minipage}
    \end{adjustbox}

    \vspace{-1.2mm}
    \begin{adjustbox}{width=0.9\linewidth,center}
        \begin{minipage}[t]{0.32\linewidth}
            \centering
            \includegraphics[width=\linewidth]{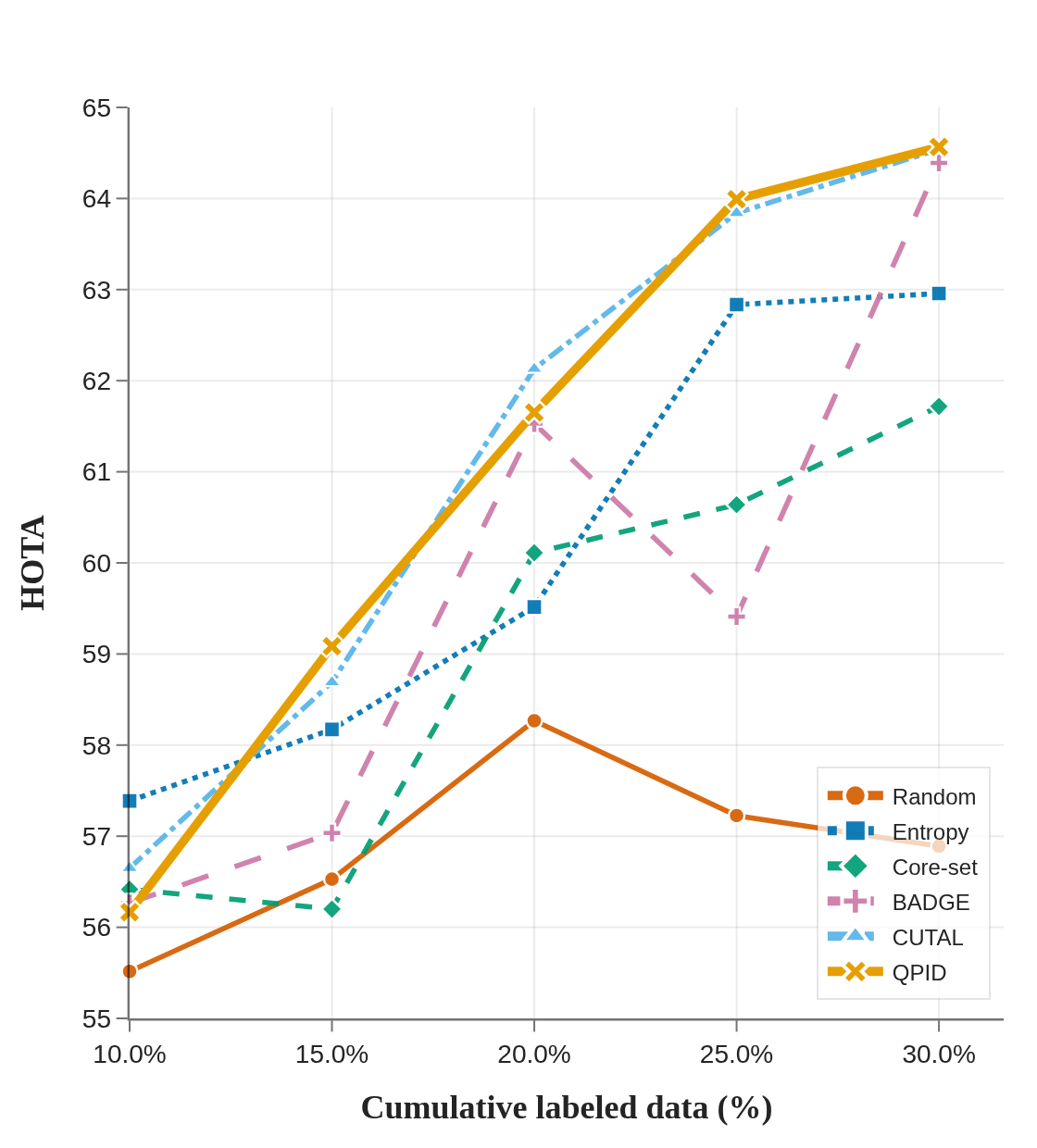}
        \end{minipage}
        \hfill
        \begin{minipage}[t]{0.32\linewidth}
            \centering
            \includegraphics[width=\linewidth]{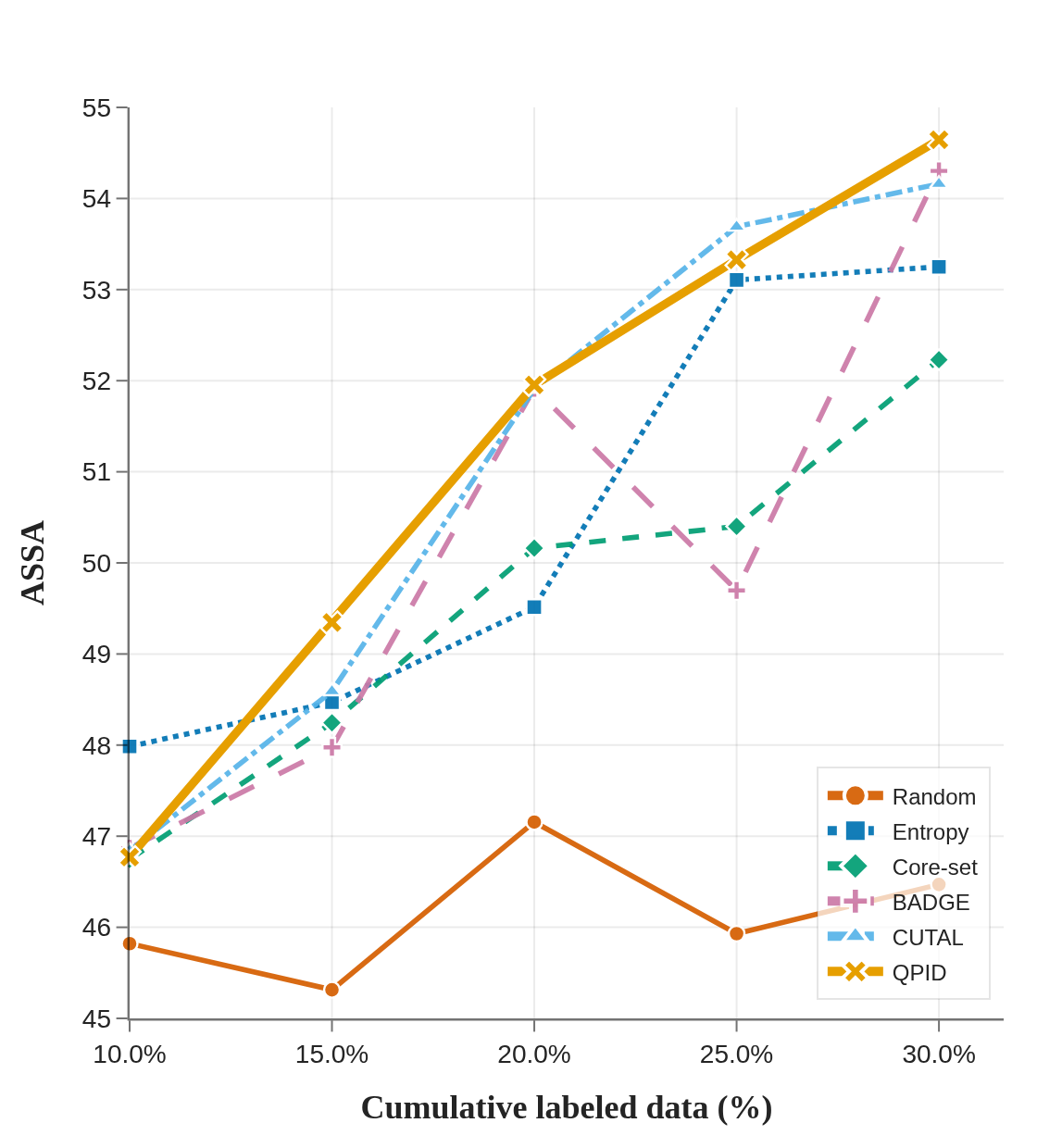}
        \end{minipage}
        \hfill
        \begin{minipage}[t]{0.32\linewidth}
            \centering
            \includegraphics[width=\linewidth]{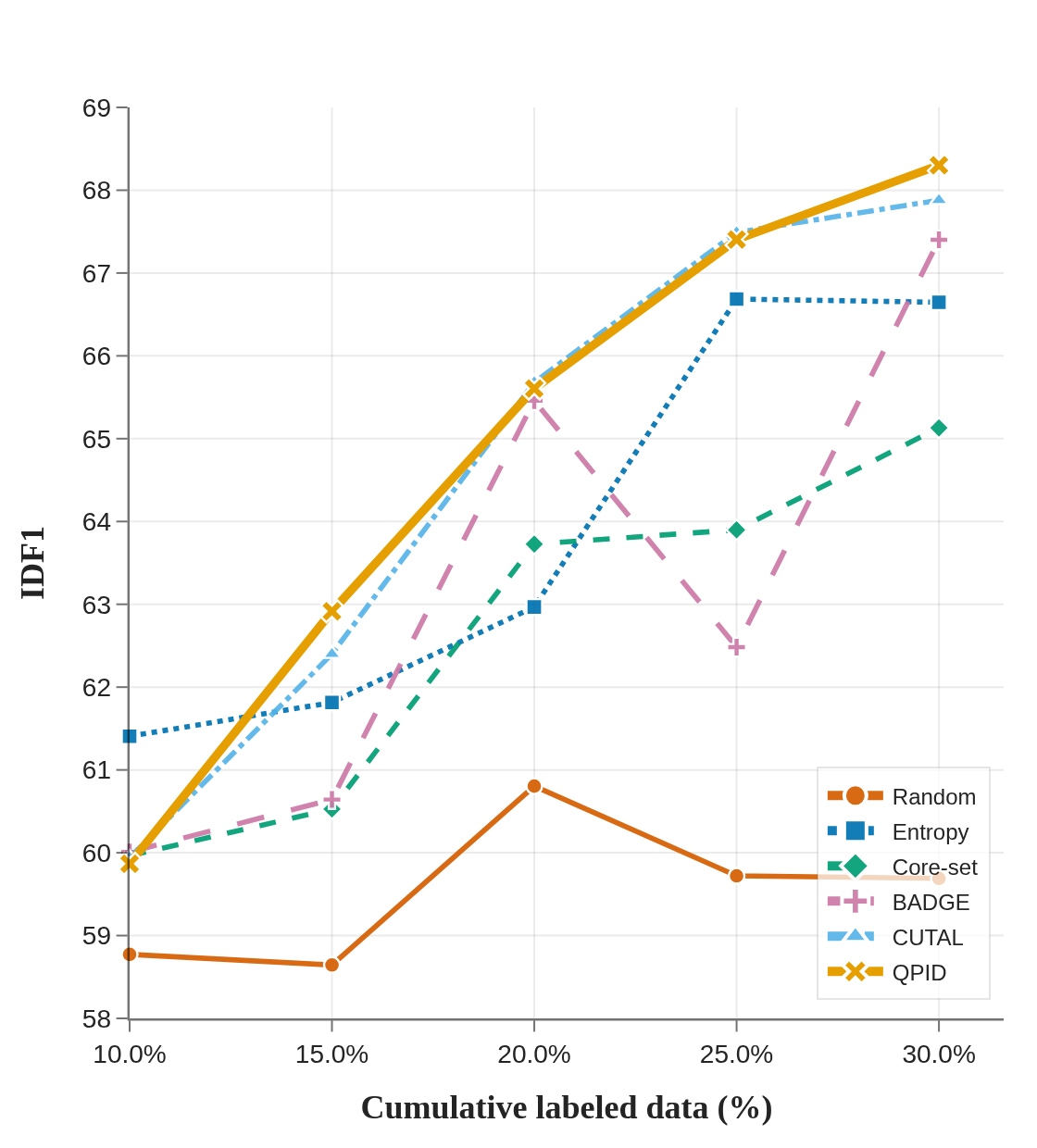}
        \end{minipage}
    \end{adjustbox}
    \vspace{-3mm}
    \caption{\textbf{Round-wise performance on SportsMOT under the 5\%+5\% schedule.}
    The upper and lower rows show results using MeMOTR and SambaMOTR, respectively.
    }
    \label{fig:memotr_samba_sportsmot_55}
\end{figure}

\vspace{-1mm}
\subsection{Datasets and Evaluation Metrics}
\label{sec:exp_datasets}
\vspace{-1mm}
We evaluate QPID on DanceTrack~\cite{sun2022dancetrack} and SportsMOT~\cite{cui2023sportsmot}.
For both datasets, the training split is used as the unlabeled pool for active learning, and all methods are evaluated on the validation split.
Official test-set evaluation servers are available, but their submission limits make them impractical for multi-round active learning, which requires repeated evaluations across rounds, methods, trackers, and seeds.
Therefore, we report validation results for all methods under the same splits, budgets, and seeds.

We use HOTA~\cite{luiten2021hota} as the primary metric and additionally report AssA~\cite{luiten2021hota} and IDF1~\cite{ristani2016performance} to assess association quality and identity consistency.
All results are averaged over three seeds.

\vspace{-1mm}
\subsection{Implementation Details}
\label{sec:exp_impl}
\vspace{-1mm}

\noindent\textbf{Trackers.}
We evaluate \textbf{MeMOTR}~\cite{gao2023memotr} and \textbf{SambaMOTR}~\cite{segu2025samba}.
The network architecture, loss functions, data augmentation, optimization, and learning-rate schedules follow the original papers and official configurations.

\noindent\textbf{Active learning protocol.}
Following the clip-level protocol of CUTAL~\cite{inoue2026clip}, we construct a clip pool $\mathcal{C}$ by sampling $T$ frames with an intra-clip interval $\Delta$ from each training video (Sec.~\ref{sec_prob_def}).
We set $T=4$ for MeMOTR and $T=10$ for SambaMOTR, with $\Delta=5$ on both datasets.
The labeling budget is defined as a percentage of the total number of training frames.
At round $r$, a frame budget $b_r$ is allocated and $B_r=\lfloor b_r/T\rfloor$ clips are selected for annotation.
We evaluate two schedules, \textbf{5\%+5\%} and \textbf{20\%+10\%}, which denote the initial labeled percentage and the active learning step size, respectively.

\noindent\textbf{Implementation of QPID.}
We set the query-embedding perturbation magnitude to $\alpha_q=0.5$ and the reference-point perturbation magnitude to $\alpha_{\tilde r}=0.01$.
For the normalization, we use $\kappa=3$ for clipping.
For Uncertainty-Weighted Visual Coverage, we set the candidate expansion ratio to \(\rho=4\).
All hyperparameters are fixed across rounds, datasets, and trackers.

\begin{table}[t]
\centering
\caption{\textbf{Round-wise results on DanceTrack under the 20\%+10\% schedule.}
Best results are marked in \textbf{bold} and second best are \underline{underlined}.}
\label{tab:dancetrack_20p10_memotr_sambamotr}
\scriptsize
\setlength{\tabcolsep}{3.5pt}
\renewcommand{\arraystretch}{1.22}

\begin{adjustbox}{width=0.75\linewidth,center}
\begin{tabular}{l|ccc|ccc|ccc}
\toprule
Method
& \multicolumn{3}{c|}{HOTA}
& \multicolumn{3}{c|}{AssA}
& \multicolumn{3}{c}{IDF1} \\
\cmidrule(lr){2-4}\cmidrule(lr){5-7}\cmidrule(lr){8-10}
& 30\% & 40\% & 50\%
& 30\% & 40\% & 50\%
& 30\% & 40\% & 50\% \\
\midrule

\multicolumn{10}{@{}l}{\textbf{MeMOTR}} \\
\addlinespace[0.15em]
Random
& 57.94 & 58.79 & 59.24
& 46.03 & 47.62 & 48.04
& 60.21 & 61.10 & 61.58 \\

Entropy
& \underline{59.21} & \underline{60.67} & 60.52
& 48.41 & \underline{49.89} & 49.86
& 61.68 & \underline{63.45} & 62.81 \\

Core-set
& 58.03 & 59.09 & 59.83
& 46.53 & 47.93 & 48.93
& 60.57 & 61.37 & 62.21 \\

BADGE
& 57.91 & 59.86 & 60.55
& 46.18 & 48.86 & 49.97
& 60.32 & 61.72 & 62.79 \\

CUTAL
& 58.87 & 60.05 & \underline{61.23}
& \underline{48.55} & 49.17 & \underline{50.60}
& \textbf{62.11} & 62.21 & \underline{63.49} \\

QPID
& \textbf{59.91} & \textbf{61.01} & \textbf{61.91}
& \textbf{49.09} & \textbf{50.54} & \textbf{52.21}
& \underline{62.03} & \textbf{63.98} & \textbf{64.79} \\

\cmidrule(lr){1-10}
\textit{Full supervision}
& \multicolumn{3}{c|}{62.63}
& \multicolumn{3}{c|}{53.12}
& \multicolumn{3}{c}{65.31} \\
\midrule

\multicolumn{10}{@{}l}{\textbf{SambaMOTR}} \\
\addlinespace[0.15em]
Random
& 48.99 & 51.69 & 50.41
& 38.91 & 41.44 & 39.99
& 52.05 & 54.68 & 53.39 \\

Entropy
& \textbf{51.51} & 51.81 & 53.48
& \underline{41.89} & \underline{42.48} & 43.33
& \underline{54.74} & 55.30 & 56.64 \\

Core-set
& 49.57 & 51.62 & 53.84
& 41.07 & 42.16 & \underline{44.91}
& 53.79 & \underline{55.86} & \underline{58.24} \\

BADGE
& 50.27 & 51.65 & 52.63
& 40.90 & 42.06 & 43.16
& 54.06 & 55.31 & 56.44 \\

CUTAL
& 50.42 & \underline{52.33} & \underline{54.08}
& 41.20 & 42.19 & 44.49
& 54.17 & 55.68 & 57.84 \\

QPID
& \underline{50.82} & \textbf{52.55} & \textbf{55.01}
& \textbf{43.19} & \textbf{43.65} & \textbf{45.79}
& \textbf{55.38} & \textbf{56.75} & \textbf{58.97} \\
\cmidrule(lr){1-10}
\textit{Full supervision}
& \multicolumn{3}{c|}{61.86}
& \multicolumn{3}{c|}{52.23}
& \multicolumn{3}{c}{65.80} \\
\bottomrule
\end{tabular}
\end{adjustbox}
\end{table}

\subsection{Baselines}
\label{sec:exp_baselines}

We compare QPID with representative random, uncertainty-based, representativeness-based, gradient-based, and clip-level MOT acquisition strategies under the same protocol.
For methods originally designed for image- or frame-level acquisition, frame-level scores or embeddings are aggregated within each clip.
Further implementation details are provided in the Supplementary Material.

\noindent\textbf{Random.}
Clips are sampled uniformly at random from the unlabeled pool.

\noindent\textbf{Entropy.}
Entropy is an uncertainty-based baseline.
For each frame, we compute the entropy of the predicted confidence scores of active tracks, take the maximum entropy over active tracks, and average the frame scores over the clip.

\noindent\textbf{Core-set~\cite{sener2017active}.}
Core-set is a representativeness-based baseline.
We construct a clip embedding by averaging decoder last-layer track-query embeddings over active tracks and sampled frames, and apply \(k\)-Center Greedy in this embedding space.

\noindent\textbf{BADGE~\cite{ash2019deep}.}
BADGE combines uncertainty and diversity through gradient embeddings.
We compute query-level gradient embeddings using the model's current predictions as hypothetical labels, aggregate them over active tracks and sampled frames to form a clip representation, and select a diverse batch via k-means++~\cite{arthur2007k}.

\noindent\textbf{CUTAL~\cite{inoue2026clip}.}
CUTAL is the most directly related clip-level active learning baseline for end-to-end MOT.
It selects clips using output-level temporal uncertainty and temporal diversity under the same clip-level protocol.

\begin{table}[t]
\centering
\caption{\textbf{Round-wise results on SportsMOT under the 20\%+10\% schedule.}
}
\label{tab:sportsmot_20p10_memotr_sambamotr}
\scriptsize
\setlength{\tabcolsep}{3.5pt}
\renewcommand{\arraystretch}{1.22}

\begin{adjustbox}{width=0.75\linewidth,center}
\begin{tabular}{l|ccc|ccc|ccc}
\toprule
Method
& \multicolumn{3}{c|}{HOTA}
& \multicolumn{3}{c|}{AssA}
& \multicolumn{3}{c}{IDF1} \\
\cmidrule(lr){2-4}\cmidrule(lr){5-7}\cmidrule(lr){8-10}
& 30\% & 40\% & 50\%
& 30\% & 40\% & 50\%
& 30\% & 40\% & 50\% \\
\midrule

\multicolumn{10}{@{}l}{\textbf{MeMOTR}} \\
\addlinespace[0.15em]
Random
& 71.81 & 73.22 & 73.77
& 61.49 & 63.28 & 64.48
& 74.48 & 75.98 & 76.71 \\

Entropy
& 71.66 & 73.33 & 74.06
& 61.17 & 63.54 & 64.86
& 74.33 & 76.34 & 77.13 \\

Core-set
& \underline{72.87} & 73.27 & 73.69
& \underline{62.80} & 63.43 & 64.34
& \underline{75.56} & 76.10 & 76.94 \\

BADGE
& 71.86 & 72.89 & 73.35
& 61.63 & 62.94 & 63.74
& 74.62 & 75.60 & 76.22 \\

CUTAL
& 72.29 & \textbf{74.22} & \underline{74.23}
& 62.28 & \textbf{65.16} & \underline{65.17}
& 75.25 & \textbf{77.67} & \underline{77.72} \\

QPID
& \textbf{73.20} & \underline{74.10} & \textbf{74.66}
& \textbf{63.47} & \underline{64.76} & \textbf{65.74}
& \textbf{76.32} & \underline{77.39} & \textbf{77.76} \\
\cmidrule(lr){1-10}
\textit{Full supervision}
& \multicolumn{3}{c|}{74.47}
& \multicolumn{3}{c|}{65.54}
& \multicolumn{3}{c}{77.47} \\
\midrule

\multicolumn{10}{@{}l}{\textbf{SambaMOTR}} \\
\addlinespace[0.15em]
Random
& 62.28 & 64.08 & 63.55
& 52.87 & 54.10 & 53.37
& 65.98 & 67.58 & 66.76 \\

Entropy
& \textbf{64.18} & 62.78 & 67.38
& \textbf{54.16} & 53.56 & \underline{58.27}
& \textbf{67.61} & 66.60 & \underline{71.29} \\

Core-set
& 61.51 & 63.39 & \underline{67.41}
& 52.99 & 53.54 & 57.46
& 65.82 & 66.86 & 70.69 \\

BADGE
& 63.97 & 65.18 & 66.43
& 53.83 & 55.25 & 56.45
& 67.16 & 68.61 & 69.67 \\

CUTAL
& 63.78 & \underline{65.94} & 66.94
& 53.67 & \underline{56.11} & 57.25
& 67.26 & \underline{69.40} & 69.99 \\

QPID
& \underline{64.13} & \textbf{66.78} & \textbf{68.64}
& \underline{54.00} & \textbf{56.63} & \textbf{59.40}
& \underline{67.38} & \textbf{70.14} & \textbf{72.39} \\
\cmidrule(lr){1-10}
\textit{Full supervision}
& \multicolumn{3}{c|}{77.27}
& \multicolumn{3}{c|}{69.51}
& \multicolumn{3}{c}{80.96} \\
\bottomrule
\end{tabular}
\end{adjustbox}
\end{table}

\subsection{Quantitative Results}
\label{sec:exp_quant}

\noindent\textbf{DanceTrack.}
Fig.~\ref{fig:memotr_samba_dancetrack_55} summarizes the round-wise learning curves under the 5\%+5\% schedule for MeMOTR (upper) and SambaMOTR (lower).
For MeMOTR, QPID performs strongly across budgets and improves association-oriented metrics, although CUTAL remains competitive in several rounds.
Under the 20\%+10\% schedule in Tab.~\ref{tab:dancetrack_20p10_memotr_sambamotr}, QPID is the best on all metrics from 40\% onward.
At 50\%, QPID outperforms the second-best method by +0.68 HOTA, +1.61 AssA, and +1.30 IDF1, narrowing the gap to full supervision to 0.72 HOTA, 0.91 AssA, and 0.52 IDF1.

For SambaMOTR, QPID is less dominant in the earliest 5\%+5\% rounds, where Entropy and CUTAL remain competitive.
Nevertheless, QPID becomes the best from 25\% onward and achieves the best final performance at 30\%, improving over the best baseline by +0.28 HOTA, +1.64 AssA, and +1.21 IDF1.
Under the 20\%+10\% schedule in Tab.~\ref{tab:dancetrack_20p10_memotr_sambamotr}, QPID is again the best on all metrics from 40\% onward.
At 50\%, it improves over the second-best method by +0.93 HOTA, +0.88 AssA, and +0.73 IDF1, while remaining below full supervision.
A post-hoc GT-based analysis further confirms that the QPID score is positively associated with clip-level association difficulty, with detailed definitions and results provided in the Supplementary Material.

\noindent\textbf{SportsMOT.}
Fig.~\ref{fig:memotr_samba_sportsmot_55} summarizes the round-wise learning curves under the 5\%+5\% schedule for MeMOTR (upper) and SambaMOTR (lower).
For MeMOTR, QPID achieves the best HOTA and IDF1 across all annotation budgets, and obtains the best AssA from 10\% to 25\% while remaining second best at 30\%.
Tab.~\ref{tab:sportsmot_20p10_memotr_sambamotr} shows that at 50\%, QPID improves over the best baseline by +0.43 HOTA, +0.57 AssA, and +0.04 IDF1, achieving performance comparable to the full-supervision reference.

For SambaMOTR, QPID is not the best in the earliest 5\%+5\% rounds, where Entropy is best at 10\% and CUTAL remains strong at 15\% and 20\%.
QPID becomes the best in HOTA from 25\% onward and achieves the best final performance at 30\%, improving over the best baseline by +0.03 HOTA, +0.35 AssA, and +0.42 IDF1.
This early-round behavior is consistent with the cold-start issue in active learning~\cite{chen2022making,chen2024making} and is further analyzed in the Supplementary Material.
Since the annotation budget is defined at the frame level, SambaMOTR's longer clips (\(T{=}10\)) yield fewer selected clips than MeMOTR's shorter clips (\(T{=}4\)) under the same budget.
This can reduce early coverage of diverse scenes and interactions, making the learned internal track states less reliable for perturbation-based acquisition.
Under the 20\%+10\% schedule in Tab.~\ref{tab:sportsmot_20p10_memotr_sambamotr}, QPID is the best on all metrics from 40\% onward for SambaMOTR.
At 50\%, it improves over the second-best method by +1.23 HOTA, +1.13 AssA, and +1.10 IDF1, showing stronger association performance than the baselines under the same frame budget.

\subsection{Ablation Study}
\label{sec:exp_ablation}

Tab.~\ref{tab:dancetrack_5p5p_components_main} ablates the two instability metrics and Uncertainty-Weighted Visual Coverage, and compares internal track-state perturbation with image-space perturbation and multiplicative with additive aggregation.
\textit{Image Gaussian Noise} adds Gaussian noise to the input images instead of perturbing internal track states.

Removing Localization Drift decreases HOTA, AssA, and IDF1 by 0.96, 1.16, and 1.03 points, respectively, while removing Entropy-Weighted Confidence Discrepancy decreases them by 0.65, 0.82, and 0.43 points.
Removing Uncertainty-Weighted Visual Coverage also reduces all three metrics, confirming the benefit of visual coverage.
Image Gaussian Noise performs substantially worse than QPID, supporting internal track-state perturbations over generic input-space noise.
Finally, additive aggregation performs worse than multiplicative aggregation, supporting the joint use of localization and confidence responses.

\begin{table}[t]
\centering
\caption{\textbf{Ablations and perturbation-space comparison on DanceTrack (MeMOTR) under the 5\%+5\% schedule.}
Results are averaged over three seeds, with Avg denoting the mean over 10\% to 30\%.}
\vspace{1mm}
\label{tab:dancetrack_5p5p_components_main}
\scriptsize
\setlength{\tabcolsep}{4.2pt}
\renewcommand{\arraystretch}{1.15}
\begin{adjustbox}{width=0.75\linewidth,center}
\begin{tabular}{l|ccc}
\toprule
Variant & HOTA (Avg) & AssA (Avg) & IDF1 (Avg) \\
\midrule
\textit{w/o Localization Drift} & 57.30 & 45.94 & 59.76 \\
\textit{w/o Entropy-Weighted Conf.\ Discrep.} & 57.61 & 46.28 & 60.36 \\
\textit{w/o Uncertainty-Weighted Visual Coverage} & 57.79 & 46.56 & 60.48 \\
\textit{Image Gaussian Noise} & 56.65 & 44.84 & 58.88 \\
\textit{Sum} & 57.70 & 46.57 & 60.38 \\
QPID & \textbf{58.26} & \textbf{47.10} & \textbf{60.79} \\
\bottomrule
\end{tabular}
\end{adjustbox}
\end{table}
\section{Conclusion}
\label{sec:conclusion}
\vspace{-1mm}

We proposed QPID, a clip acquisition method for query-propagation MOT that estimates association instability through two-sided perturbations of internal track states and selects representative high-instability clips using Uncertainty-Weighted Visual Coverage.
Experiments on DanceTrack and SportsMOT with MeMOTR and SambaMOTR demonstrate strong performance against active learning baselines under the same annotation budget.
QPID can under-rank clips in which only one instability component is large and requires \(1.18\)--\(1.19\times\) the acquisition time of CUTAL, while leaving deployment-time tracking inference unchanged.
More robust aggregation and efficient instability estimation remain directions for future work.

\bibliography{egbib}

\clearpage
\bmvaResetAuthors

\title{Supplementary Material:\\ \vspace{2mm}
Probing Association Instability with Track-State Perturbations for Clip-Level Active Learning in Query-Propagation Multi-Object Tracking}

\addauthor{Riku Inoue}{riku.inoue@ntt.com}{1}
\addauthor{Shogo Sato}{shg.sato@ntt.com}{1}
\addauthor{Kazuhiko Murasaki}{kazuhiko.murasaki@ntt.com}{1}
\addauthor{Tomoyasu Shimada}{tomoyasu.shimada@ntt.com}{1}
\addauthor{Toshihiko Nishimura}{toshihiko.nishimura@ntt.com}{1}
\addauthor{Ryuichi Tanida}{ryuichi.tanida@ntt.com}{1}

\addinstitution{
 Human Informatics Laboratories\\
 NTT, Inc.\\
 Kanagawa, Japan
}

\runninghead{INOUE ET AL.}{QPID for Clip-Level Active Learning in MOT}

\maketitle

\appendix
\section{Supplementary Overview}
\label{sec:supp_overview}

This Supplementary Material provides additional baseline implementation details, complete round-wise results under the 5\%+5\% schedule, further analyses, and qualitative visualizations.

Sec.~\ref{sec:supp_baselines} provides additional implementation details of the baselines under our clip-level protocol for query-propagation end-to-end trackers.
Sec.~\ref{sec:supp_quant} reports complete round-wise results under the 5\%+5\% schedule.
Sec.~\ref{sec:supp_ablation} presents extended ablation studies that validate key design choices in QPID.
Sec.~\ref{sec:supp_analysis} provides further analyses of association difficulty, statistical stability, hyperparameters, diversity-aware selection through Uncertainty-Weighted Visual Coverage, cold-start behavior in early rounds, the gap to full supervision in SambaMOTR~\cite{segu2025samba}, association-oriented improvements over CUTAL~\cite{inoue2026clip}, and acquisition sampling time.
Sec.~\ref{sec:supp_qual} offers qualitative visualizations of perturbation-induced prediction instability.

\section{Additional Baseline Implementation Details}
\label{sec:supp_baselines}

\noindent\textbf{Entropy.}
We compute entropy on the predicted confidence of each active track.
Active tracks are defined by a fixed confidence threshold of 0.5.
At each frame, we take the maximum entropy over active tracks, and obtain a clip score by averaging these frame scores over the sampled frames in the clip.

\noindent\textbf{Core-set~\cite{sener2017active}.}
We construct a clip embedding by averaging decoder last-layer embeddings of active track queries over tracks and sampled frames.
We then apply k-Center Greedy in this clip embedding space using cosine distance.

\noindent\textbf{BADGE~\cite{ash2019deep}.}
We adapt BADGE by constructing gradient embeddings for track queries and selecting a diverse batch via k-means++~\cite{arthur2007k}.
Pseudo labels are obtained from the predicted confidences using a fixed threshold of 0.5.
Query-level gradient embeddings are summed over active queries and sampled frames to form a clip representation, and k-means++ seeding is performed in this clip representation space.

\section{Complete Quantitative Results}
\label{sec:supp_quant}

This section reports complete round-wise results under the 5\%+5\% schedule on DanceTrack~\cite{sun2022dancetrack} and SportsMOT~\cite{cui2023sportsmot} using MeMOTR~\cite{gao2023memotr} and SambaMOTR~\cite{segu2025samba}.
We report HOTA, AssA, and IDF1, and all results are averaged over three seeds.
The corresponding results under the 20\%+10\% schedule are reported in the main paper.

Tab.~\ref{tab:dancetrack_5p5p_memotr_sambamotr_vertical} reports the complete 5\%+5\% results on DanceTrack.
Tab.~\ref{tab:sportsmot_5p5p_memotr_sambamotr_vertical} reports the complete 5\%+5\% results on SportsMOT.

\begin{table}[t]
\centering
\caption{\textbf{Round-wise results on DanceTrack under the 5\%+5\% schedule.}
Best results are marked in \textbf{bold} and second best are \underline{underlined}.}
\label{tab:dancetrack_5p5p_memotr_sambamotr_vertical}
\scriptsize
\setlength{\tabcolsep}{2.7pt}
\renewcommand{\arraystretch}{1.22}

\begin{adjustbox}{width=0.93\linewidth,center}
\begin{tabular}{l|ccccc|ccccc|ccccc}
\toprule
Method
& \multicolumn{5}{c|}{HOTA}
& \multicolumn{5}{c|}{AssA}
& \multicolumn{5}{c}{IDF1} \\
\cmidrule(lr){2-6}\cmidrule(lr){7-11}\cmidrule(lr){12-16}
& 10\% & 15\% & 20\% & 25\% & 30\%
& 10\% & 15\% & 20\% & 25\% & 30\%
& 10\% & 15\% & 20\% & 25\% & 30\% \\
\midrule

\multicolumn{16}{@{}l}{\textbf{MeMOTR}} \\
\addlinespace[0.2em]
Random
& 54.02 & 54.65 & 54.98 & 55.91 & 57.52
& 41.79 & 42.49 & 43.05 & 43.64 & 45.81
& 56.54 & 56.36 & 57.33 & 57.98 & 59.55 \\

Entropy
& 53.45 & 55.95 & 56.40 & 58.81 & 59.44
& 41.88 & 44.53 & 44.42 & 47.73 & 48.40
& 56.02 & 58.53 & 58.31 & 61.16 & 62.14 \\

Core-set
& 52.49 & 54.18 & 55.16 & 55.80 & 57.78
& 39.77 & 41.61 & 42.86 & 44.05 & 46.01
& 54.55 & 55.94 & 57.25 & 58.28 & 59.64 \\

BADGE
& 53.27 & 55.29 & 55.83 & 58.58 & 58.65
& 40.69 & 43.53 & 44.29 & 47.31 & 47.24
& 55.20 & 58.16 & 58.37 & 60.74 & 60.83 \\

CUTAL
& \underline{54.04} & \textbf{57.62} & \underline{58.61} & \underline{59.91} & \underline{60.45}
& \underline{42.03} & \underline{45.47} & \underline{47.79} & \underline{48.84} & \textbf{50.25}
& \underline{56.60} & \underline{59.44} & \underline{60.88} & \textbf{62.87} & \underline{62.75} \\

QPID
& \textbf{54.56} & \underline{57.30} & \textbf{58.90} & \textbf{60.05} & \textbf{60.51}
& \textbf{42.24} & \textbf{45.82} & \textbf{48.11} & \textbf{49.45} & \underline{49.90}
& \textbf{56.83} & \textbf{59.74} & \textbf{61.64} & \underline{62.63} & \textbf{63.13} \\

\midrule

\multicolumn{16}{@{}l}{\textbf{SambaMOTR}} \\
\addlinespace[0.2em]
Random
& 40.97 & 41.65 & 43.15 & 43.32 & 43.60
& 31.37 & 30.99 & 32.51 & 32.33 & 32.71
& 44.09 & 43.41 & 45.58 & 45.21 & 45.67 \\

Entropy
& 42.92 & \textbf{46.76} & 46.76 & 48.02 & 50.12
& 32.74 & \underline{37.48} & 37.58 & 38.40 & 39.90
& 45.77 & \textbf{50.65} & 50.55 & 51.64 & 53.11 \\

Core-set
& 42.88 & 43.39 & 42.78 & 46.81 & 46.95
& 32.72 & 34.00 & 34.89 & 37.84 & 38.48
& 46.30 & 47.48 & 47.88 & 50.48 & 51.79 \\

BADGE
& 40.82 & \underline{46.37} & 45.62 & \underline{49.28} & 48.77
& 32.03 & 36.34 & 37.91 & \underline{39.17} & 39.63
& 44.82 & 49.97 & 50.23 & \underline{53.16} & 52.49 \\

CUTAL
& \textbf{44.40} & 44.95 & \underline{48.46} & 47.13 & \underline{50.58}
& \textbf{34.78} & 35.63 & \textbf{39.57} & 38.62 & \underline{40.26}
& \textbf{47.81} & 48.74 & \textbf{52.56} & 51.70 & \underline{53.74} \\

QPID
& \underline{43.60} & 46.19 & \textbf{48.61} & \textbf{50.21} & \textbf{50.86}
& \underline{34.12} & \textbf{37.73} & \underline{38.91} & \textbf{41.96} & \textbf{41.90}
& \underline{47.21} & \underline{50.39} & \underline{52.50} & \textbf{55.09} & \textbf{54.95} \\

\bottomrule
\end{tabular}
\end{adjustbox}
\end{table}

\begin{table*}[t]
\centering
\caption{\textbf{Round-wise results on SportsMOT under the 5\%+5\% schedule.}
Best results are marked in \textbf{bold} and second best are \underline{underlined}.}
\label{tab:sportsmot_5p5p_memotr_sambamotr_vertical}
\scriptsize
\setlength{\tabcolsep}{2.7pt}
\renewcommand{\arraystretch}{1.22}

\begin{adjustbox}{width=0.93\linewidth,center}
\begin{tabular}{l|ccccc|ccccc|ccccc}
\toprule
Method
& \multicolumn{5}{c|}{HOTA}
& \multicolumn{5}{c|}{AssA}
& \multicolumn{5}{c}{IDF1} \\
\cmidrule(lr){2-6}\cmidrule(lr){7-11}\cmidrule(lr){12-16}
& 10\% & 15\% & 20\% & 25\% & 30\%
& 10\% & 15\% & 20\% & 25\% & 30\%
& 10\% & 15\% & 20\% & 25\% & 30\% \\
\midrule

\multicolumn{16}{@{}l}{\textbf{MeMOTR}} \\
\addlinespace[0.2em]
Random
& 68.38 & 69.61 & 71.60 & 71.78 & 72.31
& 57.21 & 58.54 & 61.19 & 60.98 & 62.04
& 70.65 & 70.92 & 74.41 & 74.08 & 74.83 \\

Entropy
& 68.61 & 69.95 & 70.73 & 72.24 & 72.14
& 57.50 & 58.27 & 60.93 & 62.05 & 62.29
& 71.49 & 71.94 & 73.70 & 73.70 & 75.47 \\

Core-set
& 67.52 & 68.43 & 70.78 & 72.50 & 71.95
& 56.60 & 57.03 & 60.04 & 62.25 & 61.76
& 70.67 & 70.80 & 73.19 & 75.21 & 74.42 \\

BADGE
& 68.17 & 69.78 & 71.20 & 71.02 & 72.82
& 57.22 & 58.98 & 60.78 & 60.34 & 63.13
& 70.77 & 72.26 & 73.98 & 73.63 & 76.04 \\

CUTAL
& \underline{69.35} & \underline{71.06} & \underline{71.82} & \underline{72.92} & \underline{73.56}
& \underline{57.91} & \underline{60.41} & \underline{61.70} & \underline{63.04} & \textbf{64.17}
& \underline{71.59} & \underline{73.75} & \underline{74.93} & \underline{76.02} & \underline{76.83} \\

QPID
& \textbf{69.46} & \textbf{71.41} & \textbf{72.19} & \textbf{72.96} & \textbf{73.62}
& \textbf{58.28} & \textbf{61.06} & \textbf{61.99} & \textbf{63.17} & \underline{64.10}
& \textbf{71.79} & \textbf{74.35} & \textbf{75.31} & \textbf{76.17} & \textbf{77.10} \\

\midrule

\multicolumn{16}{@{}l}{\textbf{SambaMOTR}} \\
\addlinespace[0.2em]
Random
& 55.52 & 56.53 & 58.27 & 57.23 & 56.89
& 45.82 & 45.31 & 47.16 & 45.93 & 46.47
& 58.77 & 58.65 & 60.81 & 59.72 & 59.69 \\

Entropy
& \textbf{57.39} & 58.17 & 59.52 & 62.84 & 62.96
& \textbf{47.99} & 48.47 & 49.52 & 53.11 & 53.25
& \textbf{61.40} & 61.82 & 62.97 & 66.69 & 66.65 \\

Core-set
& 56.42 & 56.20 & 60.11 & 60.64 & 61.72
& 46.75 & 48.25 & 50.16 & 50.40 & 52.23
& 59.96 & 60.53 & 63.73 & 63.90 & 65.13 \\

BADGE
& 56.27 & 57.04 & 61.53 & 59.41 & 64.39
& \underline{46.87} & 47.98 & \underline{51.92} & 49.70 & \underline{54.30}
& \underline{60.01} & 60.64 & 65.46 & 62.48 & 67.40 \\

CUTAL
& \underline{56.65} & \underline{58.69} & \textbf{62.13} & \underline{63.84} & \underline{64.54}
& 46.84 & \underline{48.59} & 51.91 & \textbf{53.69} & 54.16
& 59.94 & \underline{62.40} & \textbf{65.68} & \textbf{67.49} & \underline{67.88} \\

QPID
& 56.17 & \textbf{59.09} & \underline{61.65} & \textbf{63.99} & \textbf{64.57}
& 46.77 & \textbf{49.35} & \textbf{51.95} & \underline{53.33} & \textbf{54.65}
& 59.87 & \textbf{62.92} & \underline{65.60} & \underline{67.40} & \textbf{68.30} \\

\bottomrule
\end{tabular}
\end{adjustbox}
\end{table*}

\section{Extended Ablation Studies}
\label{sec:supp_ablation}

Tab.~\ref{tab:dancetrack_5p5p_components_supp} reports the full round-wise ablations and perturbation-space comparisons of QPID on DanceTrack with MeMOTR under the 5\%+5\% schedule, complementing the compact average-only ablation results in the main paper.
Overall, the full QPID achieves the best average performance across all three metrics, confirming the effectiveness of the full design.

Removing either instability metric degrades the averaged performance relative to the full method, indicating that Localization Drift and Entropy-Weighted Confidence Discrepancy capture complementary aspects of association instability.
The degradation is larger when removing Localization Drift across HOTA, AssA, and IDF1, indicating that perturbation-induced localization changes are especially important in this setting.

Removing Uncertainty-Weighted Visual Coverage reduces the averages to 57.79 HOTA, 46.56 AssA, and 60.48 IDF1.
Although this variant remains competitive, it is consistently below the full QPID on the averaged metrics, indicating that visual coverage improves the robustness of clip selection.

Removing scale calibration from the reference-point perturbation (\textit{w/o Scale-Calibrated Ref. Pert.}) lowers the averaged performance from 58.26 to 57.71 HOTA, from 47.10 to 46.48 AssA, and from 60.79 to 60.39 IDF1.
This result supports the importance of calibrating reference-point perturbations by the width and height of the clean propagated reference point, rather than applying uncalibrated perturbations in normalized coordinate space.

We also compare against \textit{Image Gaussian Noise}, which replaces the proposed internal track-state perturbation with image-space perturbation by adding two-sided Gaussian noise with standard deviation 0.1 to the input image while keeping the propagated clean track state unchanged across branches.
This variant performs worse than QPID on average, with 56.65 HOTA, 44.84 AssA, and 58.88 IDF1.
This gap suggests that not all perturbation-induced prediction changes are equally useful for clip acquisition in MOT.
Image-space perturbations can reflect sensitivity to local appearance or background changes, whereas QPID directly perturbs the propagated track state that is reused for maintaining identities across frames.
Therefore, the proposed internal track-state perturbation is better aligned with association difficulty than image-space perturbation in this setting.

We further compare the two-sided design with a lower-cost one-sided variant and an equal-compute variant using two independently sampled perturbation directions.
The two-sided design achieves the best averaged performance: 58.26 HOTA, 47.10 AssA, and 60.79 IDF1.
The one-sided variant obtains 57.42, 46.06, and 59.73, respectively, while the two-independent-direction variant obtains 57.73, 46.57, and 60.28.
These results empirically support the two-sided design in this setting.

Finally, replacing the multiplicative aggregation with additive aggregation (\textit{Sum}) is competitive in some individual rounds but remains inferior on average, validating the final aggregation design used in QPID.
However, multiplicative aggregation can under-rank clips in which only one instability component is large, which remains a limitation of the current formulation.

\begin{table*}[t]
\centering
\caption{\textbf{Ablations and perturbation-space comparisons of QPID on DanceTrack (MeMOTR) under the 5\%+5\% schedule.}
Best results are marked in \textbf{bold} and second best are \underline{underlined}.
Results are averaged over three seeds. Avg denotes the mean over 10\% to 30\%.
\textit{w/o Scale-Calibrated Ref. Pert.} removes scale calibration in reference-point perturbation, and
\textit{Image Gaussian Noise} replaces internal track-state perturbation with image-space perturbation.}
\label{tab:dancetrack_5p5p_components_supp}
\scriptsize
\setlength{\tabcolsep}{2.4pt}
\renewcommand{\arraystretch}{1.18}
\begin{adjustbox}{width=\linewidth,center}
\begin{tabular}{l|cccccc|cccccc|cccccc}
\toprule
Variant
& \multicolumn{6}{c|}{HOTA}
& \multicolumn{6}{c|}{AssA}
& \multicolumn{6}{c}{IDF1} \\
\cmidrule(lr){2-7}\cmidrule(lr){8-13}\cmidrule(lr){14-19}
& 10\% & 15\% & 20\% & 25\% & 30\% & Avg
& 10\% & 15\% & 20\% & 25\% & 30\% & Avg
& 10\% & 15\% & 20\% & 25\% & 30\% & Avg \\
\midrule
\textit{w/o Localization Drift}
& 53.77 & 56.15 & 57.83 & 59.05 & 59.69 & 57.30
& 41.38 & 44.70 & 46.30 & 48.28 & 49.04 & 45.94
& 56.05 & 58.45 & 60.22 & 61.49 & 62.60 & 59.76 \\
\textit{w/o Entropy-Weighted Conf.\ Discrep.}
& \textbf{54.61} & 56.20 & 57.64 & 59.43 & 60.16 & 57.61
& \textbf{42.80} & 44.33 & 46.19 & 48.48 & 49.60 & 46.28
& \textbf{57.38} & 58.69 & 60.36 & 62.56 & 62.78 & 60.36 \\
\textit{w/o Uncertainty-Weighted Visual Coverage}
& 54.26 & \textbf{57.33} & 58.60 & 59.22 & 59.55 & \underline{57.79}
& 42.32 & \textbf{45.88} & 47.58 & 48.47 & 48.52 & 46.56
& 56.57 & \textbf{60.13} & \underline{61.43} & 62.07 & 62.20 & \underline{60.48} \\
\textit{w/o Scale-Calibrated Ref. Pert.}
& 53.37 & 57.22 & 57.64 & \underline{60.01} & \underline{60.31} & 57.71
& 41.17 & 45.80 & 46.24 & \underline{49.30} & \underline{49.86} & 46.48
& 55.89 & \underline{59.94} & 60.37 & \textbf{62.76} & 62.98 & 60.39 \\
\textit{Image Gaussian Noise}
& 52.64 & 55.76 & 56.18 & 58.71 & 59.98 & 56.65
& 39.96 & 43.51 & 44.30 & 47.11 & 49.32 & 44.84
& 54.86 & 57.59 & 58.22 & 61.00 & 62.74 & 58.88 \\
\textit{One-sided}
& 54.35 & 55.59 & 58.38 & 58.49 & 60.27 & 57.42
& \underline{42.47} & 43.49 & 47.46 & 47.20 & 49.68 & 46.06
& \underline{56.94} & 57.62 & 60.81 & 60.87 & 62.41 & 59.73 \\
\textit{Two Independent Directions}
& 54.09 & 56.23 & 58.60 & 59.46 & 60.27 & 57.73
& 42.18 & 44.74 & 47.45 & 48.79 & 49.71 & \underline{46.57}
& 56.44 & 58.79 & 61.29 & 62.00 & 62.86 & 60.28 \\
\textit{Sum}
& 54.17 & 56.25 & \underline{58.83} & 59.26 & 60.00 & 57.70
& 42.15 & 44.63 & \underline{48.04} & 48.50 & 49.51 & \underline{46.57}
& 56.90 & 58.93 & 61.15 & 61.88 & \underline{63.03} & 60.38 \\
QPID
& \underline{54.56} & \underline{57.30} & \textbf{58.90} & \textbf{60.05} & \textbf{60.51} & \textbf{58.26}
& 42.24 & \underline{45.82} & \textbf{48.11} & \textbf{49.45} & \textbf{49.90} & \textbf{47.10}
& 56.83 & 59.74 & \textbf{61.64} & \underline{62.63} & \textbf{63.13} & \textbf{60.79} \\
\bottomrule
\end{tabular}
\end{adjustbox}
\end{table*}

\section{Further Analyses}
\label{sec:supp_analysis}

\subsection{Association-Difficulty Analysis}
\label{sec:supp_analysis_assoc}

To directly assess whether QPID's clip-level association-instability score \(U(c)\) reflects association difficulty, we conduct a post-hoc GT-based analysis on the round-1 unlabeled pool of DanceTrack with MeMOTR under the 5\%+5\% schedule.
Within each seed, QPID, CUTAL, and Random share the same initial labeled set, tracker checkpoint, and unlabeled pool, and GT annotations are used only for this analysis.

After matching predictions to GT at the sampled frames, we evaluate two measures of association difficulty.
The first is an AssA-style clip-local association error, \(1-\mathrm{AssA}_{\mathrm{clip}}\), which measures inconsistency between predicted track identities and GT identities across sampled frames.
The second is competing-GT IoU, defined by averaging the maximum IoU between each matched GT box and any other GT box in the same frame.
Higher values indicate greater ambiguity caused by nearby competing targets.

Across the full unlabeled pool, the Spearman correlation between the re-evaluated QPID score \(U(c)\) and the clip-local association error is \(0.659{\pm}0.007\) over three seeds.
The corresponding correlation with competing-GT IoU is \(0.501{\pm}0.026\).
The partial Spearman correlations controlling for the numbers of GT tracks and GT detections per clip remain \(0.633\) and \(0.428\), respectively, suggesting that these relationships are not explained solely by the number of targets in a clip.

Tab.~\ref{tab:gt_validation} further compares the clips selected at round 1.
QPID selects clips with higher clip-local association error and competing-GT IoU than CUTAL and Random in all three seeds.
These results provide direct evidence that higher QPID scores are associated with greater association difficulty and that QPID preferentially selects more difficult clips in this setting.

\begin{table}[t]
\centering
\caption{\textbf{GT-based analysis of round-1 selections on DanceTrack with MeMOTR under the 5\%+5\% schedule.}
Results are mean \(\pm\) standard deviation over three seeds.
Higher values indicate greater association difficulty.}
\label{tab:gt_validation}
\scriptsize
\setlength{\tabcolsep}{3.0pt}
\renewcommand{\arraystretch}{1.10}
\begin{adjustbox}{width=0.78\linewidth,center}
\begin{tabular}{lccc}
\toprule
Measure & QPID & CUTAL & Random \\
\midrule
\(1-\mathrm{AssA}_{\mathrm{clip}}\) \(\uparrow\)
& \textbf{0.223{\scriptsize\(\pm\)0.010}}
& 0.184{\scriptsize\(\pm\)0.016}
& 0.075{\scriptsize\(\pm\)0.006} \\
Competing-GT IoU \(\uparrow\)
& \textbf{0.278{\scriptsize\(\pm\)0.003}}
& 0.252{\scriptsize\(\pm\)0.003}
& 0.215{\scriptsize\(\pm\)0.005} \\
\bottomrule
\end{tabular}
\end{adjustbox}
\end{table}

\subsection{Statistical Stability}
\label{sec:supp_analysis_stability}

To assess the stability of the results across the three predefined seeds, Tabs.~\ref{tab:std_5p5p} and \ref{tab:std_20p10} report the round-wise standard deviations of QPID and CUTAL for all combinations of datasets and trackers.
The corresponding mean values under the 5\%+5\% and 20\%+10\% schedules are reported in Sec.~\ref{sec:supp_quant} and the main paper, respectively.
We focus on CUTAL as the most directly related clip-level active learning baseline.

In some settings, the performance difference between QPID and CUTAL is small relative to the variation across seeds.
For example, under the 5\%+5\% schedule with MeMOTR, the final-round AssA values are \(49.90{\pm}0.77\) for QPID and \(50.25{\pm}0.57\) for CUTAL on DanceTrack, and \(64.10{\pm}0.30\) and \(64.17{\pm}1.50\), respectively, on SportsMOT.
Across 48 matched MeMOTR comparisons, QPID yields higher HOTA, AssA, and IDF1 in 35, 30, and 32 cases, respectively.

\begin{table*}[t]
\centering
\caption{\textbf{Round-wise standard deviations under the 5\%+5\% schedule.}
Values are reported as QPID / CUTAL standard deviations over three predefined seeds.
The corresponding mean values are reported in Sec.~\ref{sec:supp_quant}.}
\label{tab:std_5p5p}
\scriptsize
\setlength{\tabcolsep}{4.0pt}
\renewcommand{\arraystretch}{1.10}
\begin{adjustbox}{width=0.9\linewidth,center}
\begin{tabular}{l|ccccc}
\toprule
Metric
& 10\% & 15\% & 20\% & 25\% & 30\% \\
\midrule

\multicolumn{6}{@{}l}{\textbf{DanceTrack with MeMOTR}} \\
HOTA
& 0.27 / 0.53
& 0.12 / 0.72
& 0.25 / 0.59
& 0.13 / 0.41
& 0.16 / 0.56 \\
AssA
& 0.42 / 0.53
& 0.44 / 0.84
& 0.57 / 1.15
& 0.17 / 0.70
& 0.77 / 0.57 \\
IDF1
& 0.83 / 0.32
& 0.52 / 0.54
& 0.50 / 0.69
& 0.46 / 0.72
& 0.18 / 1.25 \\

\midrule
\multicolumn{6}{@{}l}{\textbf{DanceTrack with SambaMOTR}} \\
HOTA
& 0.75 / 0.65
& 1.52 / 1.24
& 1.11 / 1.14
& 1.24 / 1.19
& 0.76 / 0.77 \\
AssA
& 0.74 / 0.37
& 1.86 / 0.21
& 1.46 / 2.29
& 1.81 / 0.88
& 0.62 / 1.36 \\
IDF1
& 0.60 / 0.63
& 2.05 / 0.18
& 1.68 / 1.57
& 1.52 / 0.53
& 0.88 / 1.15 \\

\midrule
\multicolumn{6}{@{}l}{\textbf{SportsMOT with MeMOTR}} \\
HOTA
& 0.12 / 0.70
& 0.92 / 1.05
& 0.47 / 0.10
& 0.45 / 0.93
& 0.28 / 0.80 \\
AssA
& 0.69 / 0.76
& 1.33 / 1.71
& 0.87 / 0.15
& 0.62 / 1.36
& 0.30 / 1.50 \\
IDF1
& 0.69 / 0.47
& 1.19 / 1.12
& 0.66 / 0.14
& 0.55 / 1.02
& 0.27 / 1.16 \\

\midrule
\multicolumn{6}{@{}l}{\textbf{SportsMOT with SambaMOTR}} \\
HOTA
& 1.59 / 1.89
& 0.29 / 0.28
& 0.65 / 1.74
& 1.50 / 0.23
& 0.28 / 1.23 \\
AssA
& 0.27 / 1.75
& 0.48 / 0.93
& 1.00 / 1.84
& 1.54 / 0.42
& 0.49 / 1.42 \\
IDF1
& 0.94 / 2.10
& 0.27 / 0.71
& 0.69 / 2.27
& 1.57 / 0.25
& 0.20 / 1.52 \\

\bottomrule
\end{tabular}
\end{adjustbox}
\end{table*}

\begin{table*}[t]
\centering
\caption{\textbf{Round-wise standard deviations under the 20\%+10\% schedule.}
Values are reported as QPID / CUTAL standard deviations over three predefined seeds.
The corresponding mean values are reported in the main paper.}
\label{tab:std_20p10}
\scriptsize
\setlength{\tabcolsep}{5.0pt}
\renewcommand{\arraystretch}{1.10}
\begin{adjustbox}{width=0.65\linewidth,center}
\begin{tabular}{l|ccc}
\toprule
Metric
& 30\% & 40\% & 50\% \\
\midrule

\multicolumn{4}{@{}l}{\textbf{DanceTrack with MeMOTR}} \\
HOTA
& 0.64 / 0.57
& 0.45 / 0.76
& 0.25 / 0.43 \\
AssA
& 1.08 / 0.92
& 0.52 / 1.30
& 0.41 / 0.77 \\
IDF1
& 1.39 / 0.94
& 0.43 / 1.01
& 0.73 / 0.92 \\

\midrule
\multicolumn{4}{@{}l}{\textbf{DanceTrack with SambaMOTR}} \\
HOTA
& 1.73 / 0.46
& 1.75 / 0.54
& 1.08 / 1.20 \\
AssA
& 1.39 / 0.25
& 1.93 / 0.86
& 1.35 / 1.96 \\
IDF1
& 1.68 / 0.27
& 1.55 / 0.76
& 1.22 / 1.19 \\

\midrule
\multicolumn{4}{@{}l}{\textbf{SportsMOT with MeMOTR}} \\
HOTA
& 0.42 / 0.32
& 0.54 / 0.24
& 0.35 / 0.45 \\
AssA
& 0.59 / 0.47
& 0.48 / 0.53
& 0.59 / 0.57 \\
IDF1
& 0.57 / 0.47
& 0.33 / 0.37
& 0.48 / 0.57 \\

\midrule
\multicolumn{4}{@{}l}{\textbf{SportsMOT with SambaMOTR}} \\
HOTA
& 1.09 / 1.49
& 0.55 / 0.50
& 1.01 / 0.31 \\
AssA
& 1.66 / 1.90
& 0.63 / 0.54
& 1.69 / 0.35 \\
IDF1
& 1.67 / 1.85
& 0.84 / 0.63
& 1.29 / 0.21 \\

\bottomrule
\end{tabular}
\end{adjustbox}
\end{table*}

\subsection{Hyperparameter Analysis}
\label{sec:supp_analysis_hyper}

Tabs.~\ref{tab:dancetrack_5p5p_alpha_q}, \ref{tab:dancetrack_5p5p_alpha_r}, \ref{tab:dancetrack_5p5p_kappa}, and \ref{tab:dancetrack_5p5p_rho} study the effect of the main hyperparameters of QPID on DanceTrack with MeMOTR
under the 5\%+5\% schedule.
These hyperparameters include the query-embedding perturbation magnitude $\alpha_q$, the reference-point perturbation magnitude $\alpha_{\tilde r}$, the normalization clipping range $\kappa$, and the candidate expansion ratio $\rho$.

For $\alpha_q$, the default value of $0.50$ achieves the best average performance across all three metrics.
Using a smaller value of $0.30$ or a larger value of $0.70$ remains competitive, but both yield slightly lower averages.

For $\alpha_{\tilde r}$, the default value of $0.010$ achieves the best average performance across all three metrics.
The smaller value $0.005$ performs best at 10\%, and the larger value $0.050$ is competitive in some individual rounds, but neither setting matches the default in the overall averages.

For $\kappa$, the default value of $3$ gives the best average performance across all three metrics.
The setting $\kappa=1$ is the second-best choice in the averaged results, whereas $\kappa=5$ leads to lower averages.
Overall, these results support $\kappa=3$ as a stable default choice for normalization.

For $\rho$, the default value of $4$ gives the best average performance across all three metrics.
A smaller value of $2$ restricts the candidate pool too strongly, whereas a larger value of $8$ remains competitive but slightly lowers the averaged metrics.
These results suggest that $\rho=4$ provides a balanced candidate pool size, maintaining focus on high-instability clips while allowing sufficient visual coverage.

Overall, these results show that QPID is not highly sensitive to moderate hyperparameter variations.
Across the evaluated settings, the default configuration remains the best or near-best choice overall, supporting its use as a stable setting in the main experiments.

\begin{table}[t]
\centering
\caption{\textbf{Hyperparameter analysis of $\alpha_q$ on DanceTrack (MeMOTR) under the 5\%+5\% schedule.}
Best results are marked in \textbf{bold} and second best are \underline{underlined}.
Results are averaged over three seeds. Avg denotes the mean over 10\% to 30\%.}
\label{tab:dancetrack_5p5p_alpha_q}
\scriptsize
\setlength{\tabcolsep}{2.7pt}
\renewcommand{\arraystretch}{1.22}
\begin{adjustbox}{width=\linewidth,center}
\begin{tabular}{l|cccccc|cccccc|cccccc}
\toprule
$\alpha_q$
& \multicolumn{6}{c|}{HOTA}
& \multicolumn{6}{c|}{AssA}
& \multicolumn{6}{c}{IDF1} \\
\cmidrule(lr){2-7}\cmidrule(lr){8-13}\cmidrule(lr){14-19}
& 10\% & 15\% & 20\% & 25\% & 30\% & Avg
& 10\% & 15\% & 20\% & 25\% & 30\% & Avg
& 10\% & 15\% & 20\% & 25\% & 30\% & Avg \\
\midrule
0.30
& \textbf{54.57} & 55.99 & 58.52 & 59.06 & \underline{60.14} & 57.66
& \textbf{42.26} & 44.03 & \underline{47.52} & 47.97 & \underline{49.48} & 46.25
& 56.65 & 58.15 & 61.09 & 61.38 & \underline{62.98} & 60.05 \\
0.50 (\textit{default})
& \underline{54.56} & \textbf{57.30} & \textbf{58.90} & \textbf{60.05} & \textbf{60.51} & \textbf{58.26}
& \underline{42.24} & \textbf{45.82} & \textbf{48.11} & \textbf{49.45} & \textbf{49.90} & \textbf{47.10}
& \textbf{56.83} & \textbf{59.74} & \textbf{61.64} & \textbf{62.63} & \textbf{63.13} & \textbf{60.79} \\
0.70
& 53.96 & \underline{56.65} & \underline{58.54} & \underline{59.70} & 59.83 & \underline{57.73}
& 41.80 & \underline{44.86} & 47.50 & \underline{48.86} & 48.91 & \underline{46.39}
& \underline{56.72} & \underline{59.16} & \underline{61.27} & \underline{62.21} & 62.23 & \underline{60.32} \\
\bottomrule
\end{tabular}
\end{adjustbox}
\end{table}

\begin{table}[t]
\centering
\caption{\textbf{Hyperparameter analysis of $\alpha_{\tilde r}$ on DanceTrack (MeMOTR) under the 5\%+5\% schedule.}
Best results are marked in \textbf{bold} and second best are \underline{underlined}.
Results are averaged over three seeds. Avg denotes the mean over 10\% to 30\%.}
\label{tab:dancetrack_5p5p_alpha_r}
\scriptsize
\setlength{\tabcolsep}{2.7pt}
\renewcommand{\arraystretch}{1.22}
\begin{adjustbox}{width=\linewidth,center}
\begin{tabular}{l|cccccc|cccccc|cccccc}
\toprule
$\alpha_{\tilde r}$
& \multicolumn{6}{c|}{HOTA}
& \multicolumn{6}{c|}{AssA}
& \multicolumn{6}{c}{IDF1} \\
\cmidrule(lr){2-7}\cmidrule(lr){8-13}\cmidrule(lr){14-19}
& 10\% & 15\% & 20\% & 25\% & 30\% & Avg
& 10\% & 15\% & 20\% & 25\% & 30\% & Avg
& 10\% & 15\% & 20\% & 25\% & 30\% & Avg \\
\midrule
0.005
& \textbf{55.01} & 55.58 & \underline{58.11} & 59.70 & \underline{59.85} & \underline{57.65}
& \textbf{42.73} & 43.29 & \underline{47.02} & 48.80 & \underline{49.05} & 46.18
& \textbf{57.35} & 57.74 & \underline{60.44} & 62.17 & \underline{62.57} & 60.05 \\
0.010 (\textit{default})
& \underline{54.56} & \textbf{57.30} & \textbf{58.90} & \textbf{60.05} & \textbf{60.51} & \textbf{58.26}
& \underline{42.24} & \textbf{45.82} & \textbf{48.11} & \textbf{49.45} & \textbf{49.90} & \textbf{47.10}
& \underline{56.83} & \underline{59.74} & \textbf{61.64} & \textbf{62.63} & \textbf{63.13} & \textbf{60.79} \\
0.050
& 54.37 & \underline{56.95} & 57.96 & \underline{59.76} & 58.93 & 57.59
& 42.21 & \underline{45.50} & 46.59 & \underline{48.88} & 47.81 & \underline{46.20}
& 56.71 & \textbf{59.76} & 60.24 & \underline{62.38} & 61.44 & \underline{60.10} \\
\bottomrule
\end{tabular}
\end{adjustbox}
\end{table}

\begin{table}[t]
\centering
\caption{\textbf{Hyperparameter analysis of $\kappa$ on DanceTrack (MeMOTR) under the 5\%+5\% schedule.}
Best results are marked in \textbf{bold} and second best are \underline{underlined}.
Results are averaged over three seeds. Avg denotes the mean over 10\% to 30\%.}
\label{tab:dancetrack_5p5p_kappa}
\scriptsize
\setlength{\tabcolsep}{2.7pt}
\renewcommand{\arraystretch}{1.22}
\begin{adjustbox}{width=\linewidth,center}
\begin{tabular}{l|cccccc|cccccc|cccccc}
\toprule
$\kappa$
& \multicolumn{6}{c|}{HOTA}
& \multicolumn{6}{c|}{AssA}
& \multicolumn{6}{c}{IDF1} \\
\cmidrule(lr){2-7}\cmidrule(lr){8-13}\cmidrule(lr){14-19}
& 10\% & 15\% & 20\% & 25\% & 30\% & Avg
& 10\% & 15\% & 20\% & 25\% & 30\% & Avg
& 10\% & 15\% & 20\% & 25\% & 30\% & Avg \\
\midrule
1
& \underline{54.06} & 56.41 & \underline{58.44} & \textbf{60.27} & 59.32 & \underline{57.70}
& \underline{41.83} & 44.74 & \underline{47.52} & \textbf{49.83} & 48.02 & \underline{46.39}
& \underline{56.41} & 59.05 & \underline{61.37} & \textbf{63.20} & 61.98 & \underline{60.40} \\
3 (\textit{default})
& \textbf{54.56} & \textbf{57.30} & \textbf{58.90} & \underline{60.05} & \textbf{60.51} & \textbf{58.26}
& \textbf{42.24} & \textbf{45.82} & \textbf{48.11} & \underline{49.45} & \textbf{49.90} & \textbf{47.10}
& \textbf{56.83} & \textbf{59.74} & \textbf{61.64} & \underline{62.63} & \textbf{63.13} & \textbf{60.79} \\
5
& 53.53 & \underline{56.82} & 57.67 & 59.66 & \underline{59.74} & 57.48
& 41.14 & \underline{45.21} & 46.52 & 49.20 & \underline{48.83} & 46.18
& 55.38 & \underline{59.53} & 60.53 & 62.48 & \underline{62.28} & 60.04 \\
\bottomrule
\end{tabular}
\end{adjustbox}
\end{table}

\begin{table}[t]
\centering
\caption{\textbf{Hyperparameter analysis of $\rho$ on DanceTrack (MeMOTR) under the 5\%+5\% schedule.}
Best results are marked in \textbf{bold} and second best are \underline{underlined}.
Results are averaged over three seeds. Avg denotes the mean over 10\% to 30\%.}
\label{tab:dancetrack_5p5p_rho}
\scriptsize
\setlength{\tabcolsep}{2.7pt}
\renewcommand{\arraystretch}{1.22}
\begin{adjustbox}{width=\linewidth,center}
\begin{tabular}{l|cccccc|cccccc|cccccc}
\toprule
$\rho$
& \multicolumn{6}{c|}{HOTA}
& \multicolumn{6}{c|}{AssA}
& \multicolumn{6}{c}{IDF1} \\
\cmidrule(lr){2-7}\cmidrule(lr){8-13}\cmidrule(lr){14-19}
& 10\% & 15\% & 20\% & 25\% & 30\% & Avg
& 10\% & 15\% & 20\% & 25\% & 30\% & Avg
& 10\% & 15\% & 20\% & 25\% & 30\% & Avg \\
\midrule
2
& 53.71 & \underline{56.79} & 57.57 & 58.80 & \underline{59.79} & 57.33
& 41.37 & \underline{45.28} & 46.09 & 47.56 & \underline{48.95} & 45.85
& 55.66 & \underline{59.39} & 60.09 & 61.41 & \underline{62.61} & 59.83 \\

4 (\textit{default})
& \underline{54.56} & \textbf{57.30} & \textbf{58.90} & \underline{60.05} & \textbf{60.51} & \textbf{58.26}
& \underline{42.24} & \textbf{45.82} & \textbf{48.11} & \underline{49.45} & \textbf{49.90} & \textbf{47.10}
& \underline{56.83} & \textbf{59.74} & \textbf{61.64} & \underline{62.63} & \textbf{63.13} & \textbf{60.79} \\

8
& \textbf{54.59} & 56.75 & \underline{58.15} & \textbf{60.39} & 59.46 & \underline{57.87}
& \textbf{42.46} & 44.95 & \underline{47.14} & \textbf{49.90} & 48.67 & \underline{46.62}
& \textbf{56.94} & 59.31 & \underline{60.91} & \textbf{63.62} & 61.98 & \underline{60.55} \\
\bottomrule
\end{tabular}
\end{adjustbox}
\end{table}

\subsection{Effect of Diversity-Aware Selection}
\label{sec:supp_analysis_diversity}

\begin{figure}[t]
    \centering
    \begin{minipage}[t]{0.32\linewidth}
        \centering
        \includegraphics[width=\linewidth]{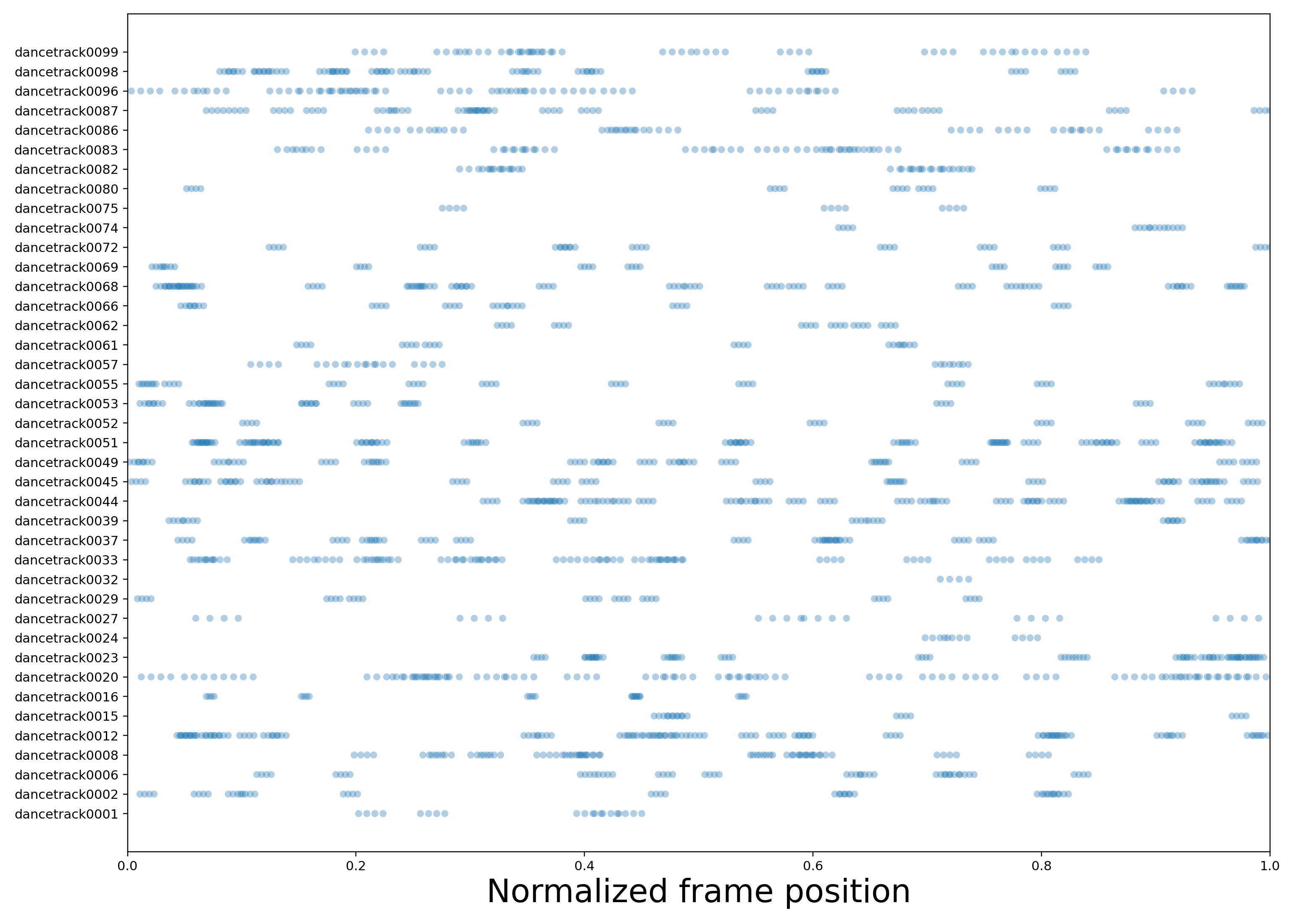}
    \end{minipage}
    \hfill
    \begin{minipage}[t]{0.32\linewidth}
        \centering
        \includegraphics[width=\linewidth]{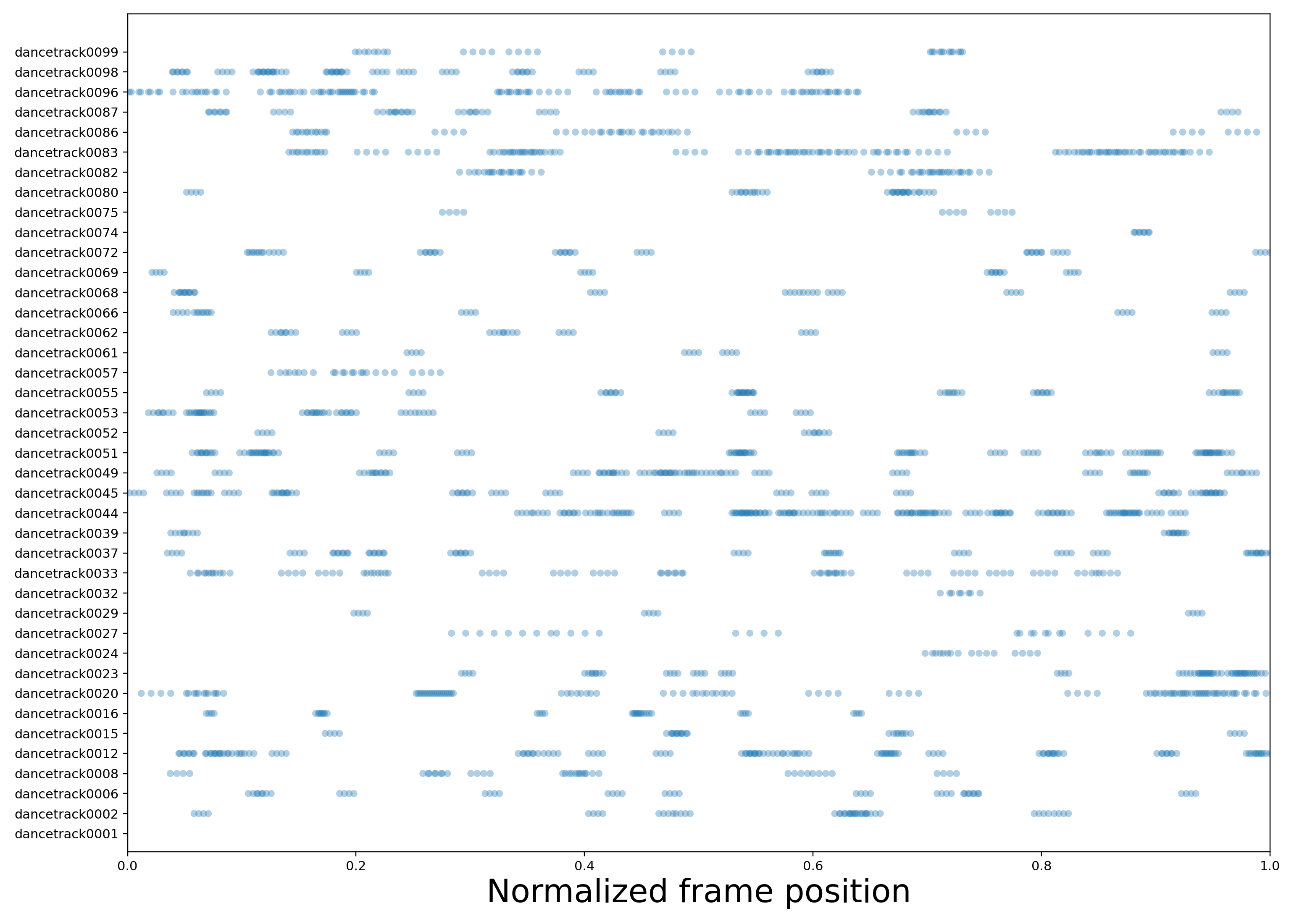}
    \end{minipage}
    \hfill
    \begin{minipage}[t]{0.32\linewidth}
        \centering
        \includegraphics[width=\linewidth]{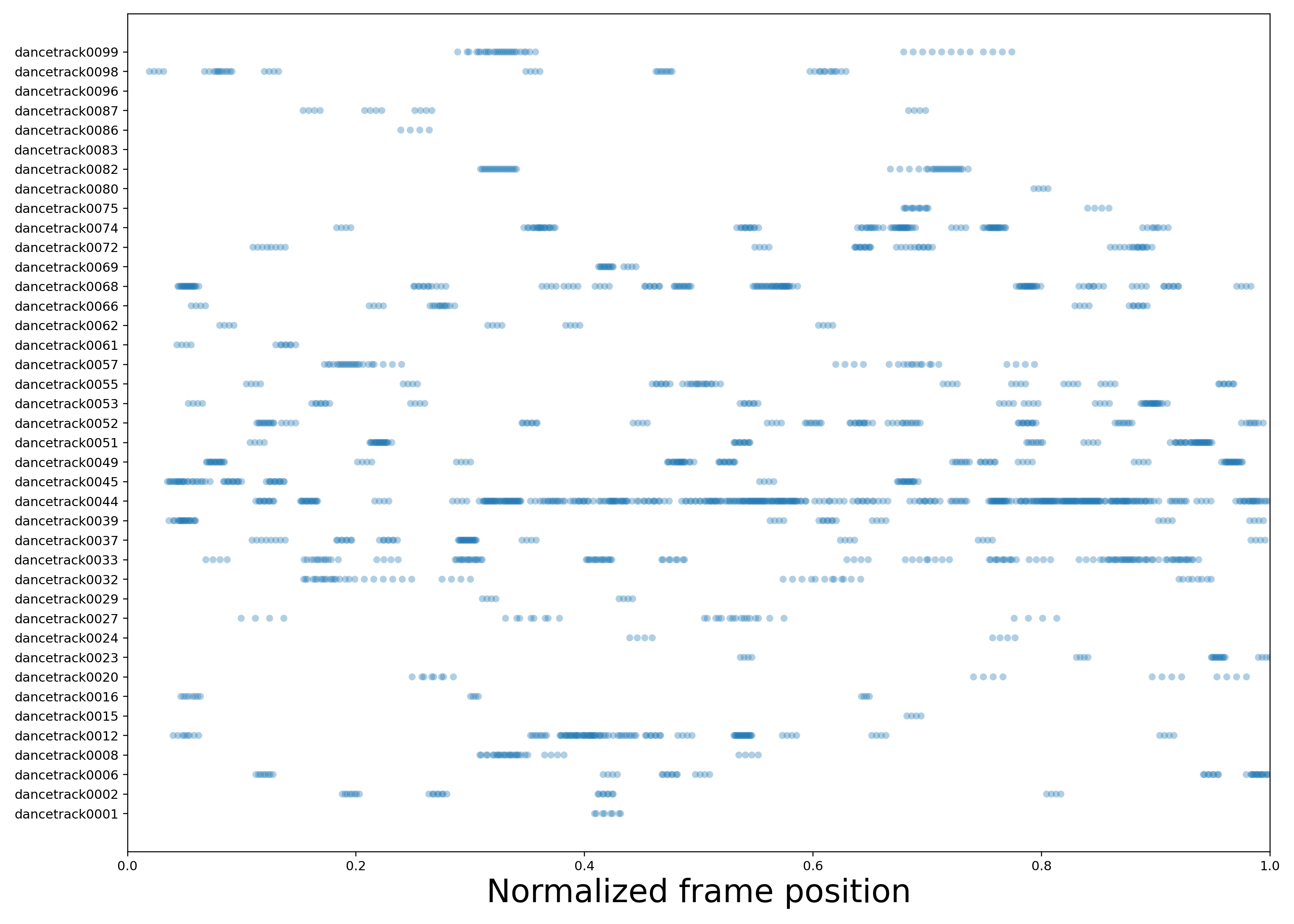}
    \end{minipage}
    \caption{\textbf{Visualization of selected frame positions on DanceTrack at Round 1 under the 5\%+5\% schedule.}
    From left to right, we show QPID, \textit{w/o Uncertainty-Weighted Visual Coverage}, and Entropy.
    Each dot denotes one selected frame.
    The horizontal axis shows the normalized temporal position within each sequence, and the vertical axis shows the sequence identity.
    QPID shows less redundant concentration than Entropy and the variant without Uncertainty-Weighted Visual Coverage.}
    \label{fig:selection_position_dancetrack}
\end{figure}

Fig.~\ref{fig:selection_position_dancetrack} visualizes the selected frame positions on DanceTrack at Round 1 under the 5\%+5\% schedule.
We compare QPID, \textit{w/o Uncertainty-Weighted Visual Coverage}, and Entropy.
Each dot denotes one selected frame, where the horizontal axis indicates the normalized temporal position within a sequence and the vertical axis indicates the sequence identity.

Compared with Entropy, QPID avoids overly concentrated selection on a small number of local temporal regions while still covering a broader range of temporal positions and sequences.
Removing Uncertainty-Weighted Visual Coverage leads to more redundant selections within similar temporal neighborhoods and sequences.
These visualizations are consistent with the role of Uncertainty-Weighted Visual Coverage in reducing redundant concentration during selection.

\subsection{Cold-Start Behavior in Early Rounds}
\label{sec:supp_analysis_coldstart}

QPID is already competitive in the early rounds with MeMOTR and generally outperforms the baselines as the labeling budget increases.
In contrast, with SambaMOTR, baselines can occasionally outperform QPID in the earliest rounds.
For example, on SportsMOT under the 5\%+5\% schedule in Tab.~\ref{tab:sportsmot_5p5p_memotr_sambamotr_vertical}, Entropy achieves the best HOTA, AssA, and IDF1 at 10\%, while CUTAL~\cite{inoue2026clip} achieves the best HOTA and IDF1 at 20\%. QPID is nevertheless the best method for HOTA from 25\% onward and for IDF1 at 30\%.
Under the 20\%+10\% schedule reported in the main paper, Entropy remains competitive at the first round of 30\% for SambaMOTR, especially in HOTA, whereas QPID becomes the best on all metrics from 40\% onward.

We attribute this trend mainly to the cold-start regime in active learning~\cite{chen2022making,chen2024making} and the longer clip unit used by SambaMOTR.
In the earliest rounds, the tracker is trained on only a small randomly sampled initial labeled subset, so the resulting supervision is limited.
This effect is more pronounced in SambaMOTR because its longer clips ($T{=}10$) reduce the number of uniquely labeled clips acquired under the same frame budget and also introduce more temporally redundant observations within each selected clip.
As a result, the learned representations and internal track states remain less reliable in the early rounds, making it harder to identify truly informative clips for acquisition.
Because QPID relies on perturbation-induced instability of internal track states, such immature states can make the estimated instability less aligned with the true association difficulty of a clip.
In this regime, simpler heuristics such as Entropy can remain competitive.

Nevertheless, this disadvantage is mainly limited to the earliest rounds.
As more labeled clips are accumulated, the tracker states become more reliable and the perturbation-based instability measured by QPID better reflects genuine association ambiguity.
Accordingly, QPID becomes increasingly effective in later rounds and eventually surpasses the baselines for SambaMOTR, while remaining strong throughout for MeMOTR.

\subsection{Full-Supervision Gap in SambaMOTR}
\label{sec:supp_samba_gap}

QPID with MeMOTR approaches full-supervision performance, whereas a substantial gap still remains for SambaMOTR.
Under the 20\%+10\% schedule reported in the main paper, the gap between QPID and full supervision is small for MeMOTR on DanceTrack, where QPID is 0.72 HOTA, 0.91 AssA, and 0.52 IDF1 below full supervision at 50\%.
On SportsMOT, QPID with MeMOTR achieves performance comparable to full supervision in this setting.
In contrast, the gap remains much larger for SambaMOTR.
At 50\%, QPID is still 6.85 HOTA, 6.44 AssA, and 6.83 IDF1 below full supervision on DanceTrack, and 8.63 HOTA, 10.11 AssA, and 8.57 IDF1 below full supervision on SportsMOT.

This gap is likely related to the larger sampling unit used by SambaMOTR.
Because SambaMOTR uses longer clips ($T{=}10$) than MeMOTR ($T{=}4$), the same frame-level budget yields substantially fewer uniquely labeled clips.
This reduces the effective coverage of diverse scenes, motion patterns, and interaction cases that can be acquired during active learning.
Moreover, each selected clip spans a longer temporal range and therefore tends to include more temporally redundant observations, so a larger fraction of the annotation budget is spent on frames with limited marginal training value.

As a result, even when the acquisition strategy becomes more reliable in later rounds, the overall supervision available to SambaMOTR remains less diverse than that of MeMOTR under the same frame-level budget.
Nevertheless, in the later rounds, QPID brings SambaMOTR closer to the full-supervision upper bound than the competing baselines, indicating that association-aware acquisition remains beneficial even in this setting.

\subsection{Association-Oriented Improvements over CUTAL}
\label{sec:supp_analysis_association_delta}

QPID is designed to select clips that are informative for improving identity association in query-propagation MOT.
We therefore further compare QPID with CUTAL, the most directly related clip-level active learning baseline, using association-oriented metrics.
For AssA and IDF1, we report QPID minus CUTAL, where positive values indicate improvement.

Tab.~\ref{tab:cutal_qpid_association_delta} summarizes the differences under both active learning schedules.
For the 5\%+5\% schedule, Final denotes the 30\% annotation budget and Avg denotes the mean over 10\%, 15\%, 20\%, 25\%, and 30\%.
For the 20\%+10\% schedule, Final denotes the 50\% annotation budget and Avg denotes the mean over 30\%, 40\%, and 50\%.

Under the 5\%+5\% schedule, QPID improves CUTAL by +0.46 AssA and +0.47 IDF1 on average across dataset--tracker settings.
Under the 20\%+10\% schedule, the gains become larger, with +1.05 AssA and +0.88 IDF1 on average.
These results indicate that perturbation-based association-instability scoring selects clips that are more useful for improving identity association than CUTAL's output-level temporal uncertainty and temporal diversity.

\begin{table}[t]
\centering
\caption{
\textbf{Association-metric differences over CUTAL under both active learning schedules.}
We report QPID minus CUTAL.
For the 5\%+5\% schedule, Final denotes 30\% and Avg denotes the mean over 10\%, 15\%, 20\%, 25\%, and 30\%.
For the 20\%+10\% schedule, Final denotes 50\% and Avg denotes the mean over 30\%, 40\%, and 50\%.
Positive values indicate improvement.
}
\label{tab:cutal_qpid_association_delta}
\scriptsize
\setlength{\tabcolsep}{3.0pt}
\renewcommand{\arraystretch}{1.18}
\begin{adjustbox}{width=0.9\linewidth,center}
\begin{tabular}{lll|rrrr}
\toprule
Schedule & Dataset & Tracker
& $\Delta$AssA Final
& $\Delta$AssA Avg
& $\Delta$IDF1 Final
& $\Delta$IDF1 Avg \\
\midrule

5\%+5\%
& DanceTrack & MeMOTR
& -0.35 & +0.23
& +0.38 & +0.29 \\

5\%+5\%
& DanceTrack & SambaMOTR
& +1.64 & +1.15
& +1.21 & +1.12 \\

5\%+5\%
& SportsMOT & MeMOTR
& -0.07 & +0.27
& +0.27 & +0.32 \\

5\%+5\%
& SportsMOT & SambaMOTR
& +0.49 & +0.17
& +0.42 & +0.14 \\

\addlinespace[0.2em]
5\%+5\%
& \textit{Average} & \textit{All settings}
& \textbf{+0.43} & \textbf{+0.46}
& \textbf{+0.57} & \textbf{+0.47} \\

\midrule

20\%+10\%
& DanceTrack & MeMOTR
& +1.61 & +1.17
& +1.30 & +1.00 \\

20\%+10\%
& DanceTrack & SambaMOTR
& +1.30 & +1.58
& +1.13 & +1.14 \\

20\%+10\%
& SportsMOT & MeMOTR
& +0.57 & +0.45
& +0.04 & +0.28 \\

20\%+10\%
& SportsMOT & SambaMOTR
& +2.15 & +1.00
& +2.40 & +1.09 \\

\addlinespace[0.2em]
20\%+10\%
& \textit{Average} & \textit{All settings}
& \textbf{+1.41} & \textbf{+1.05}
& \textbf{+1.22} & \textbf{+0.88} \\

\bottomrule
\end{tabular}
\end{adjustbox}
\end{table}

\subsection{Acquisition Sampling Time}
\label{sec:supp_analysis_overhead}

QPID introduces additional computation only during active learning acquisition.
It does not modify the model architecture, training objective, or deployment-time tracking inference.
To quantify this cost, we measure the running time of a single full Round 1 acquisition sampling with MeMOTR.
CUTAL and QPID are measured under the same hardware, batch size, data-loading configuration, and unlabeled pool for each dataset.

As shown in Tab.~\ref{tab:round1_sampling_time}, QPID takes 321.95 minutes on DanceTrack and 214.75 minutes on SportsMOT, compared with 273.73 and 180.58 minutes for CUTAL, respectively.
This corresponds to \(1.18\) to \(1.19\times\) the acquisition sampling time of CUTAL.
This overhead is relatively modest for annotation acquisition, because it is incurred only at active learning rounds and does not affect deployment-time tracking speed.

\begin{table}[t]
\centering
\caption{
\textbf{Acquisition sampling time comparison.}
We report the running time of a single full Round 1 acquisition sampling command with MeMOTR, including model initialization, acquisition scoring, batch selection, and output.
}
\label{tab:round1_sampling_time}
\scriptsize
\setlength{\tabcolsep}{4.0pt}
\renewcommand{\arraystretch}{1.18}
\begin{adjustbox}{width=0.8\linewidth,center}
\begin{tabular}{llrr}
\toprule
Dataset & Method & Time [min] $\downarrow$ & Rel. to CUTAL $\downarrow$ \\
\midrule
DanceTrack & CUTAL & 273.73 & 1.00$\times$ \\
DanceTrack & QPID & 321.95 & 1.18$\times$ \\
\midrule
SportsMOT & CUTAL & 180.58 & 1.00$\times$ \\
SportsMOT & QPID & 214.75 & 1.19$\times$ \\
\bottomrule
\end{tabular}
\end{adjustbox}
\end{table}

\section{Qualitative Results}
\label{sec:supp_qual}

\begin{figure}[t]
    \centering
    
    \begin{tabular}{@{}c c@{}}
        \includegraphics[width=0.48\linewidth]{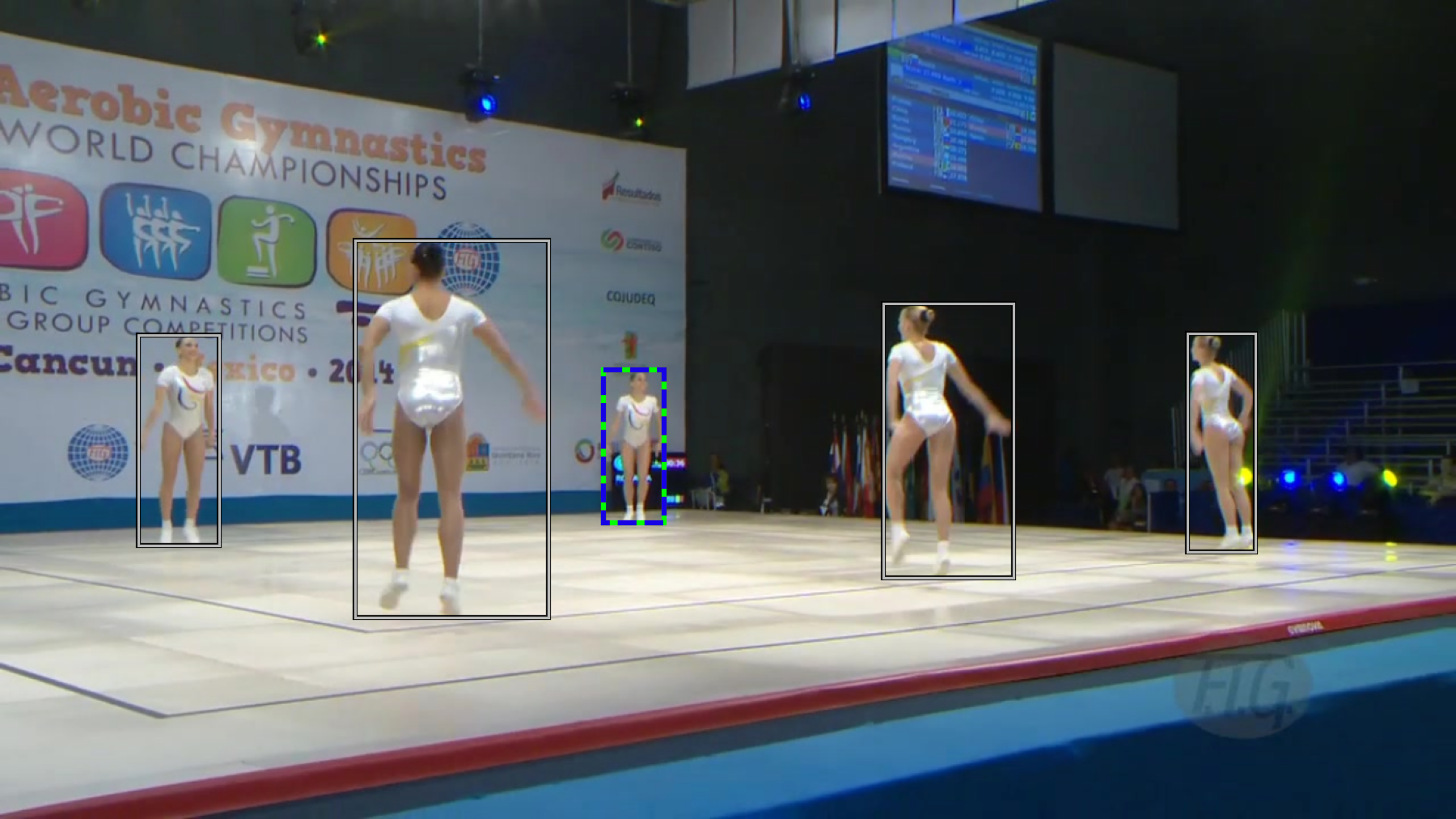} &
        \includegraphics[width=0.48\linewidth]{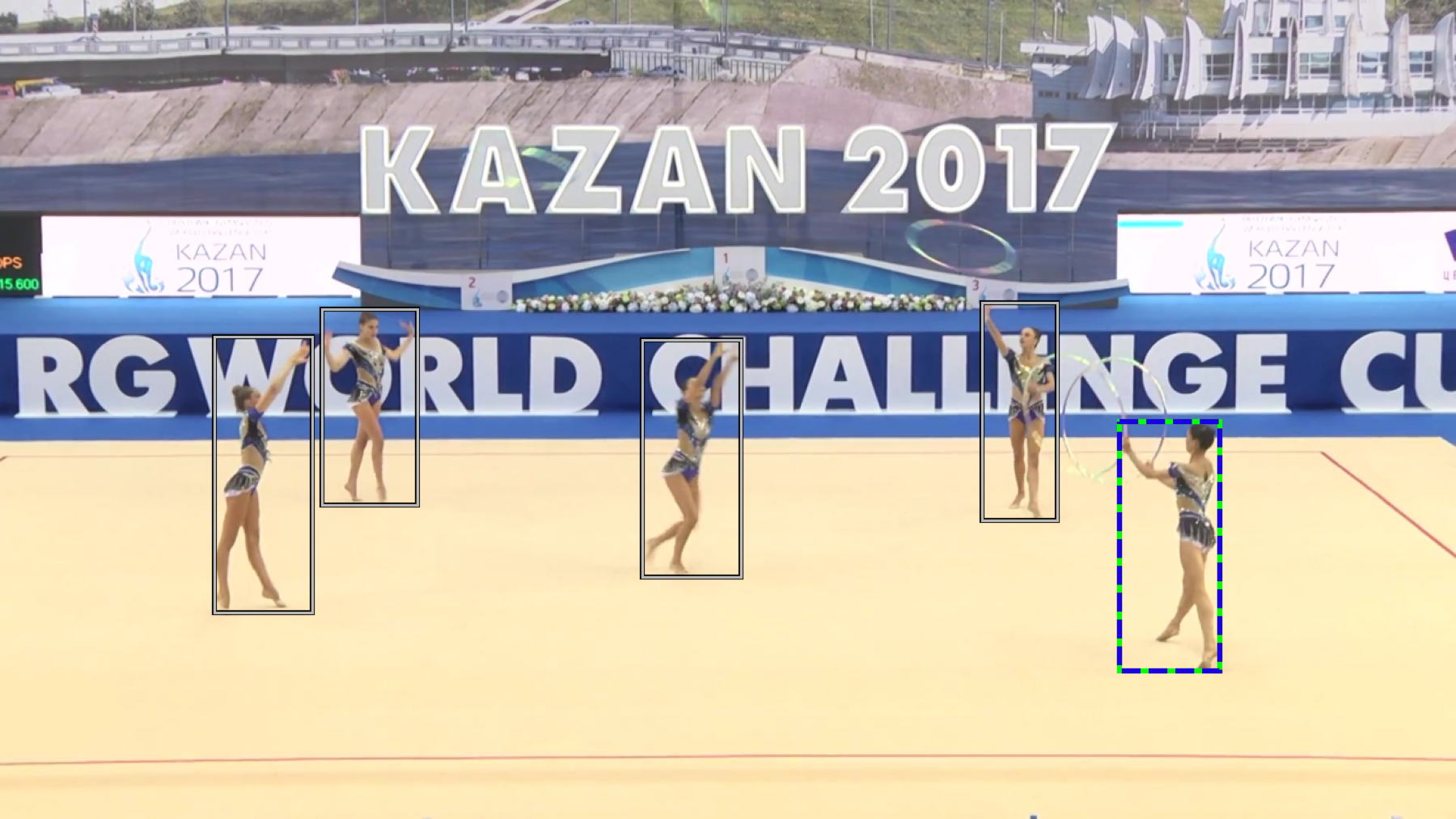} \\
        \includegraphics[width=0.48\linewidth]{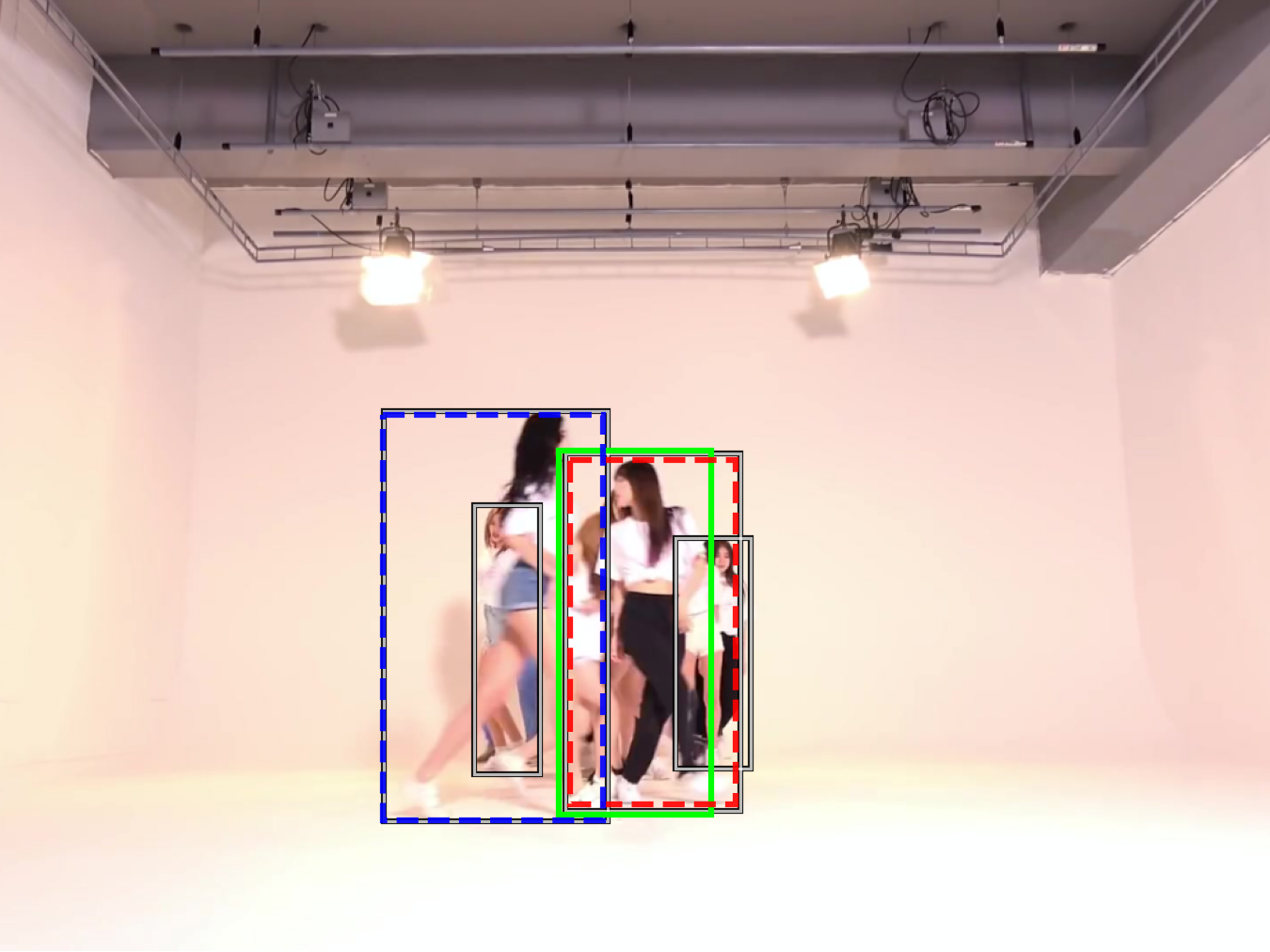} &
        \includegraphics[width=0.48\linewidth]{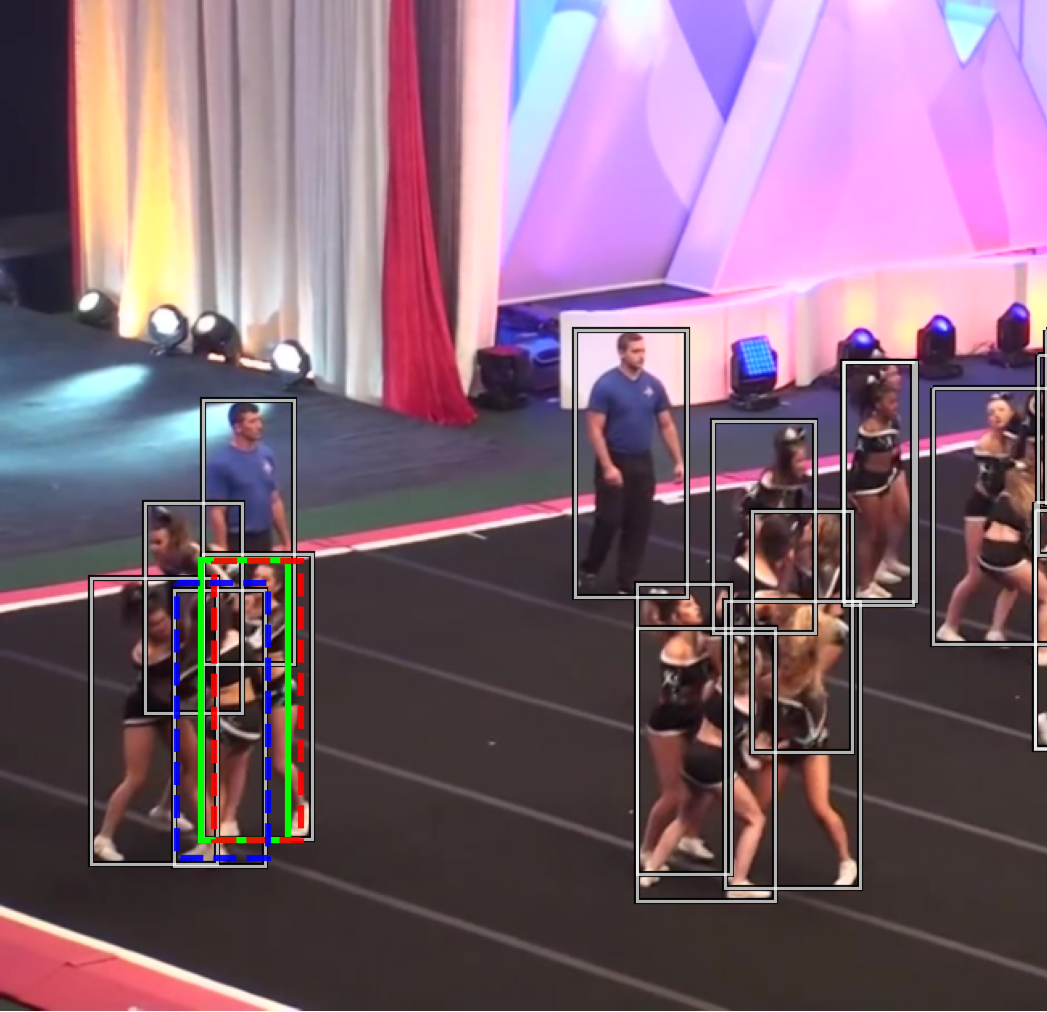}
    \end{tabular}

    \caption{\textbf{Qualitative comparison of stable and unstable clips under perturbations on DanceTrack.}
    The top row shows clips with low $U(c)$ and the bottom row shows clips with high $U(c)$.
    Green, blue, and red boxes denote the clean prediction and the two perturbed predictions for the most unstable object.}
    \label{fig:qual_res_dancetrack}
\end{figure}

\begin{figure}[t]
    \centering
    
    \begin{tabular}{@{}c c@{}}
        \includegraphics[width=0.48\linewidth]{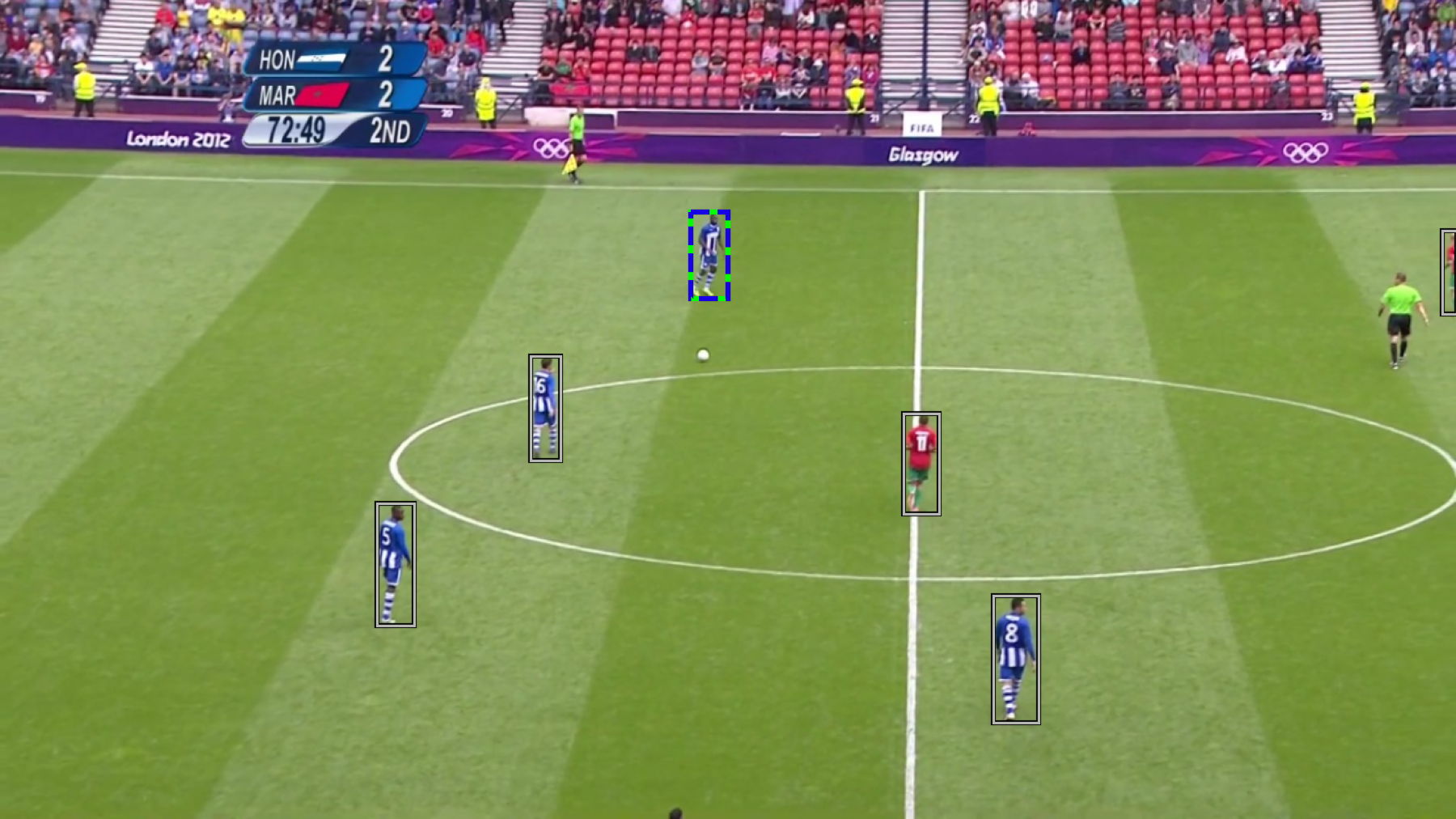} &
        \includegraphics[width=0.48\linewidth]{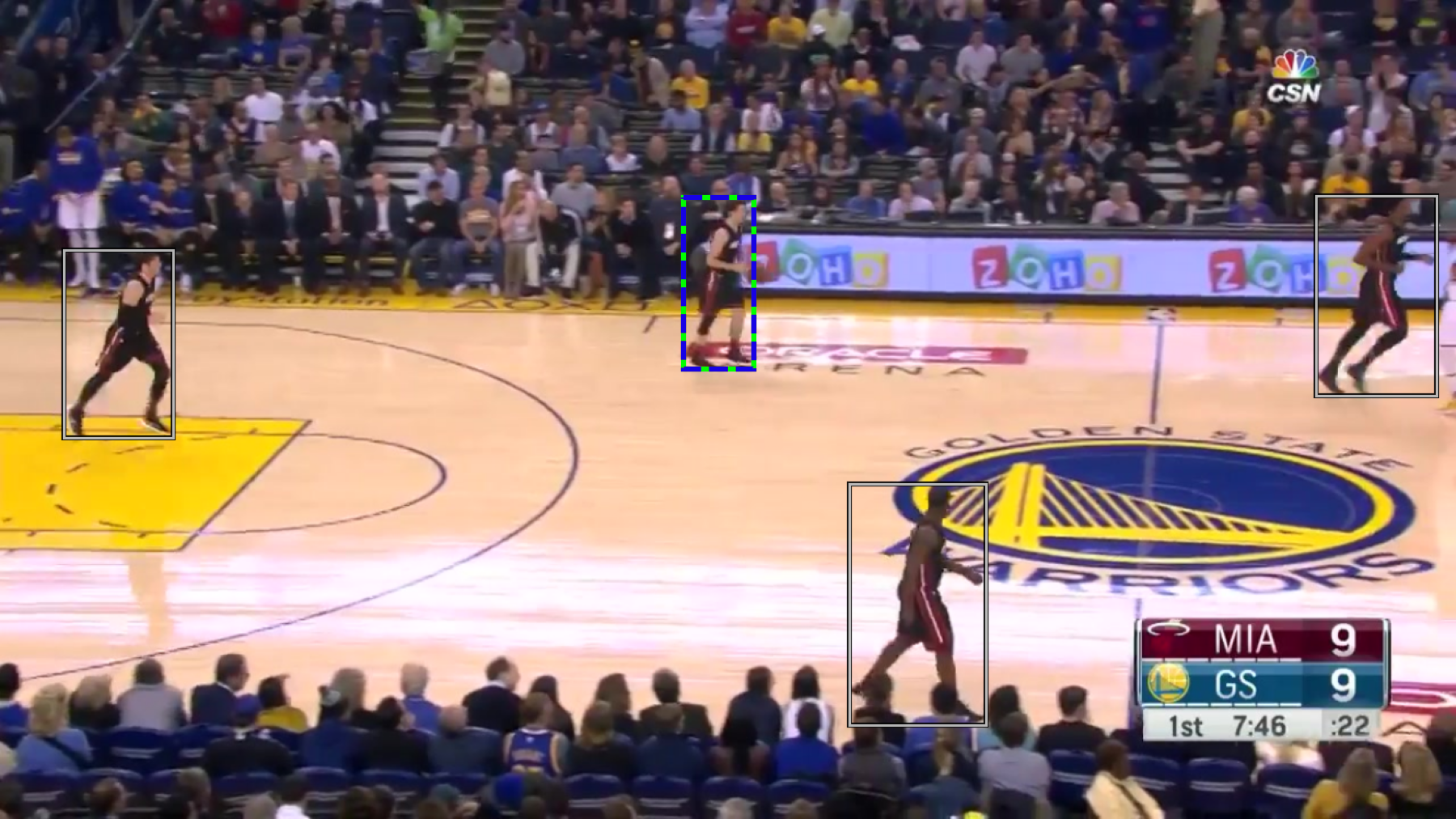} \\
        \includegraphics[width=0.48\linewidth]{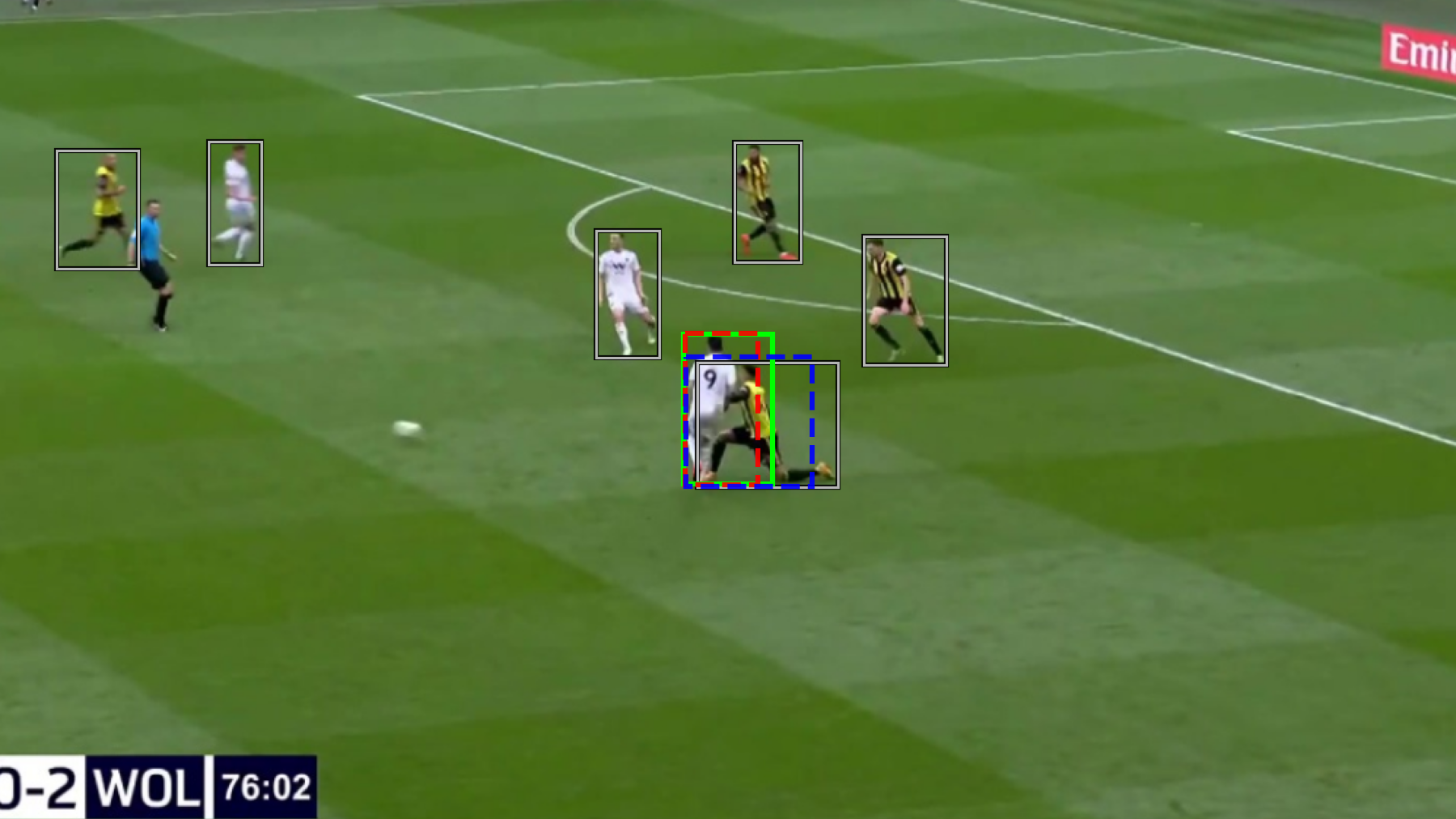} &
        \includegraphics[width=0.48\linewidth]{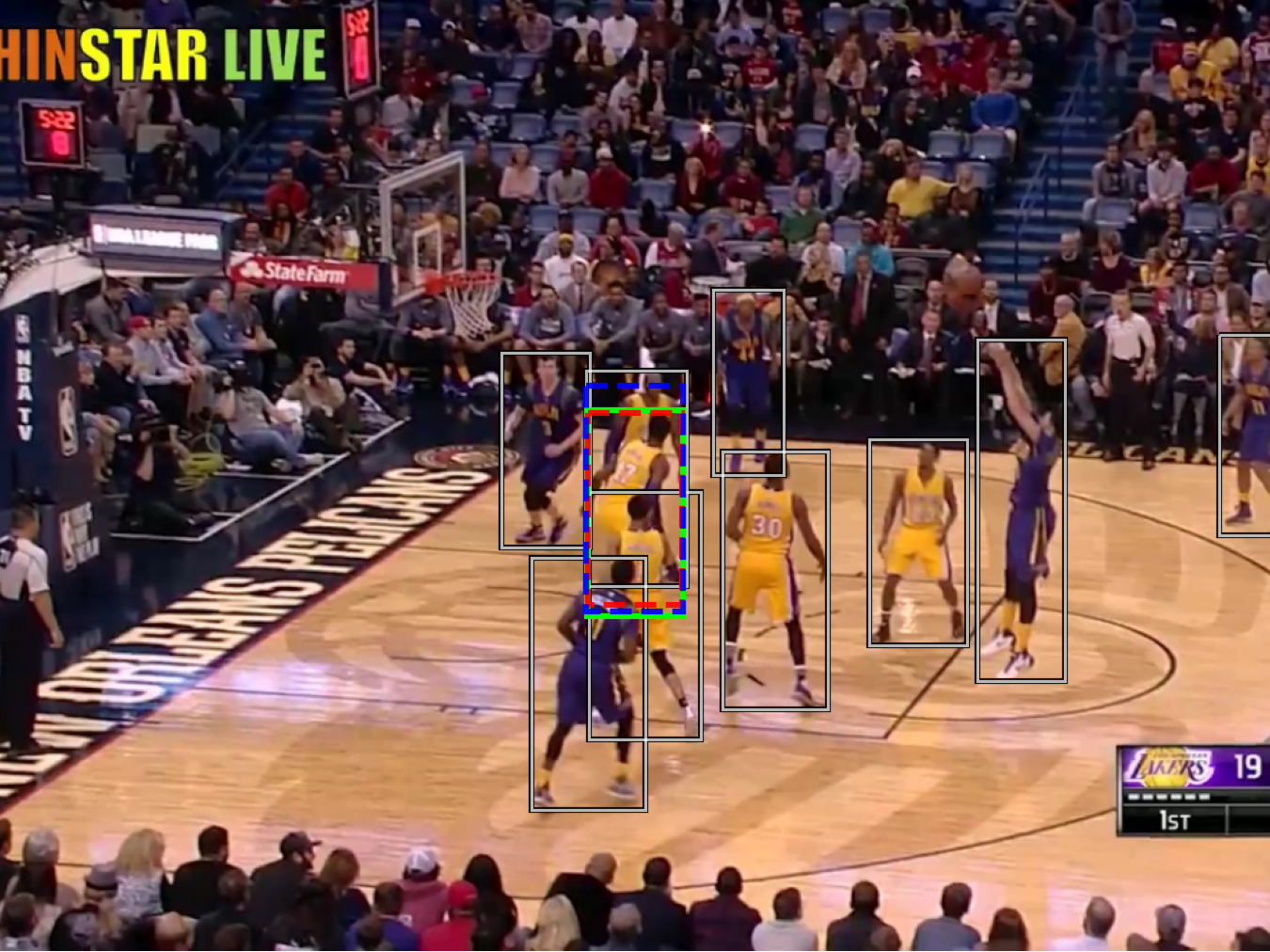}
    \end{tabular}

    \caption{\textbf{Qualitative comparison of stable and unstable clips under perturbations on SportsMOT.}
    The top row shows clips with low $U(c)$ and the bottom row shows clips with high $U(c)$.
    Green, blue, and red boxes denote the clean prediction and the two perturbed predictions for the most unstable object.}
    \label{fig:qual_res_sportsmot}
\end{figure}

Figs.~\ref{fig:qual_res_dancetrack} and \ref{fig:qual_res_sportsmot} show qualitative examples of perturbation-induced variations in predicted bounding boxes on DanceTrack and SportsMOT, respectively.
All examples are obtained using models trained with 20\% labeled frames under the 20\%+10\% schedule.
Within each dataset, the top row shows stable clips with low $U(c)$, whereas the bottom row shows unstable clips with high $U(c)$ selected by QPID.

In each example, the green, blue, and red boxes correspond to the clean prediction and the two perturbed predictions for the most unstable object in the scene, while the remaining boxes show the standard inference outputs.
Stable examples exhibit only small variations under perturbations and often correspond to sparse scenes with limited interactions.
By contrast, unstable examples show larger perturbation-induced box shifts and typically involve dense crowds or visually similar nearby objects, which increase association ambiguity and lead to higher prediction instability.

\end{document}